\documentclass[lettersize,journal]{IEEEtran}
\usepackage{amsmath,amsfonts}
\usepackage{algorithmic}
\usepackage{algorithm}
\usepackage{array}
\usepackage[caption=false,font=normalsize,labelfont=sf,textfont=sf]{subfig}
\usepackage{textcomp}
\usepackage{stfloats}
\usepackage{url}
\usepackage{verbatim}
\usepackage{graphicx}
\usepackage{cite}
\usepackage{tabularx}
\usepackage{threeparttable}
\usepackage[figuresright]{rotating}
\usepackage{multirow} 
\usepackage{color}
\usepackage{gensymb}
\usepackage[section]{placeins}
\usepackage{float}
\usepackage{makecell}

\begin{document}

\title{A Survey on Monocular Re-Localization: \\
        From the Perspective of Scene Map Representation}

\author{Jinyu Miao$^{1}$, Kun Jiang$^{1,*}$, Tuopu Wen$^{1,*}$, Yunlong Wang$^{1}$, Peijin Jia$^{1}$, Xuhe Zhao$^{1}$, \\
Qian Cheng$^{1}$, Zhongyang Xiao$^{2}$, Jin Huang$^{1}$, Zhihua Zhong$^{1}$, Diange Yang$^{1,*}$ 
\thanks{This work was supported in part by the National Natural Science Foundation of China under Grants U22A20104 and 52372414, and Beijing Municipal Science and Technology Commission (Grant No.Z221100008122011).}
\thanks{$^{1}$Jinyu Miao, Kun Jiang, Tuopu Wen, Yunlong Wang, Peijin Jia, Xuhe Zhao, Qian Cheng, Jin Huang, Zhihua Zhong, and Diange Yang are with the School of Vehicle and Mobility, Tsinghua University, Beijing, 100084, P. R. China.{\tt\small jinyu.miao97@gmail.com}}%
\thanks{$^{2}$Zhongyang Xiao is with Autonomous Driving Division of NIO Inc., Beijing, P. R. China}%
\thanks{$^{*}$Corresponding author: Diange Yang, Kun Jiang, and Tuopu Wen}
}

\markboth{Journal of \LaTeX\ Class Files,~Vol.~14, No.~8, August~2021}%
{Shell \MakeLowercase{\textit{et al.}}: A Sample Article Using IEEEtran.cls for IEEE Journals}


\maketitle

\begin{abstract}
    Monocular Re-Localization (MRL) is a critical component in autonomous applications, estimating 6 degree-of-freedom ego poses \textit{w.r.t.} the scene map based on monocular images.
    In recent decades, significant progress has been made in the development of MRL techniques. Numerous algorithms have accomplished extraordinary success in terms of localization accuracy and robustness.
    In MRL, scene maps are represented in various forms, and they determine how MRL methods work and how MRL methods perform.
    However, to the best of our knowledge, existing surveys do not provide systematic reviews about the relationship between MRL solutions and their used scene map representation.
    This survey fills the gap by comprehensively reviewing MRL methods from such a perspective, promoting further research.
    1) We commence by delving into the problem definition of MRL, exploring current challenges, and comparing ours with existing surveys.
    2) Many well-known MRL methods are categorized and reviewed into five classes according to the representation forms of utilized map, \textit{i.e.}, geo-tagged frames, visual landmarks, point clouds, vectorized semantic map, and neural network-based map.
    3) To quantitatively and fairly compare MRL methods with various map, we introduce some public datasets and provide the performances of some state-of-the-art MRL methods. The strengths and weakness of MRL methods with different map are analyzed.
    4) We finally introduce some topics of interest in this field and give personal opinions.
    This survey can serve as a valuable referenced materials for MRL, and a continuously updated summary of this survey is publicly available to the community at:
    \begin{center}
        \texttt{\url{github.com/jinyummiao/map-in-mono-reloc}}
    \end{center}
\end{abstract}


\section{Introduction}
\label{sec:intro}

\IEEEPARstart{V}{isual} Re-Localization, as a typical state estimation problem, is a challenging and ongoing topic.
Using a low-cost monocular camera, Monocular Re-Localization (MRL) tends to estimate 6 Degree-of-Freedom (DoF) poses, including orientation and position, with regards to scene map when the vehicle re-visits a mapped area \cite{VBL2018PR,LCD2022TITS}.
This task holds great attentions in various autonomous applications, \textit{e.g.}, Virtual Reality (VR) \cite{Sven2014ECCV,Jonathan2014TVCG}, robots navigation \cite{Lim2015IJRR,LATITUDE,LocNeRF}, Autonomous Driving (AD) \cite{TM3Loc,MLVHM,DA4AD,RoadMap}. Generally, MRL is commonly discussed to solve long-term association (so called loop closure) problem in Simultaneous Localization and Mapping (SLAM) system \cite{LCD2022TITS}, recover ego pose as a state re-initialization step in kidnapped robot scenarios \cite{Sotirios2020CVC}, or retrieve prior information from a pre-built map to complement online perception systems \cite{nmp}.

\begin{figure}[!t]
    \centering
    \includegraphics[width=0.97\linewidth]{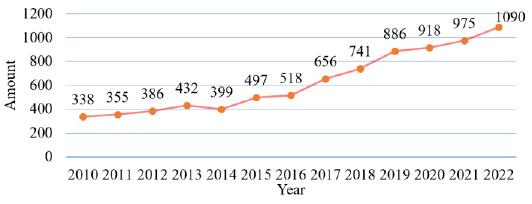}
    \caption{An overview of the development in MRL. a) published papers titled by ``visual re-localization'' in each year, b) some of landmarks in MRL researches.}
    \label{fig:intro}
\end{figure}

As the spatial model of the explored scene, map saves prior knowledge of scene (such as appearance, geometry, topology, \textit{etc.}) and serves a reference coordinate for MRL.
Generally, the MRL methods align currently observed sensory data to scene map to solve ego pose, which can be summarized as: \textbf{build scene map in a specific representation form (mapping stage)} and then \textbf{localize based on the matching information between current observations and the map (localization stage)}.
In MRL solution, the data in the map is captured by sensors in the historical period, and it probably has extreme differences in visual conditions (like weather \cite{seqslam}, illumination \cite{vprbench}, day-night change \cite{Asha2019ICRA}, \textit{etc.}) and even modality \cite{mono_i2p_vo,cmrnet} with the currently observed data, making MRL a technically hard problem.
Therefore, the representation form of data in scene map plays a vital role in MRL task and affects the robustness of MRL methods against real-world interference.

\begin{figure*}[!t]
    \centering
    \includegraphics[width=0.97\linewidth]{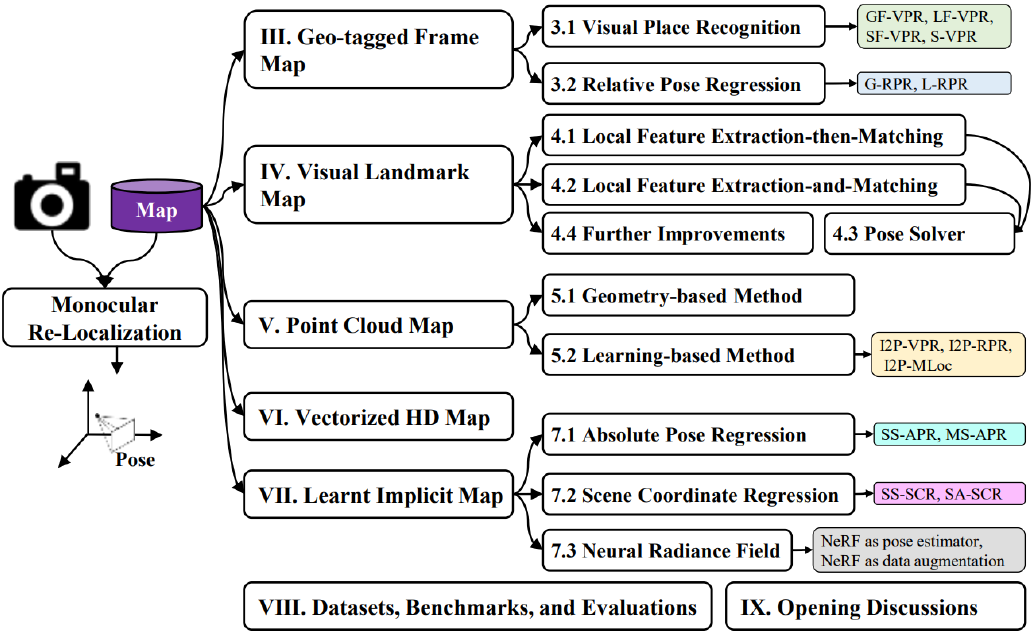}
    \caption{The structure of this survey.}
    \label{fig:structure}
\end{figure*}

Various kinds of map have been involved in the existing MRL researches to achieve robust, efficient, and high-precision localization.
In some applications without 3-dimensional (3D) scene model, map is represented by frames with geographic location annotation (named as \textbf{geo-tagged frame map}) so that localization can be achieved by image retrieval \cite{netvlad,patch-netvlad} or relative pose estimation \cite{arnold2022mapfree,RelocNet}. For most of MRL methods tending to high localization accuracy, map is composed of visual point clouds with high-dimensional descriptors (named as \textbf{visual landmark map}) \cite{hloc1,hloc2}. In this formulation, the alignment can be performed by pixel-wise matching between current observations and map data. Some cross-modal MRL methods adopt LiDAR \textbf{point cloud map} \cite{cmrnet,mono_i2p_vo} for its high geometry accuracy and density. In the area of AD, the compact and vectorized semantic map, called \textbf{vectorized High-Definition (HD) map}, is widely utilized \cite{TM3Loc,MLVHM}. So the alignment process becomes instance-wise matching between semantic map elements. To achieve improvement by the benefit of End-to-End (E2E) scheme as other computer vision task \cite{alexnet,faster-rcnn}, some methods learn to extract pose-related information from implicit map and localize itself, which no longer require explicit scene map \cite{posenet,neumap}. We believe that the scene maps of these methods are implicitly represented in a form of network parameters and name such a kind of map as \textbf{learnt implicit map}. In this study, we find that MRL methods with the same kind of scene map usually perform in a theoretically similar way: \textbf{the map representation determines the implementation way and even the localization performance of MRL methods}.

However, existing surveys in related field do not discuss about the relationship between MRL methods and the scene map, which blocking the analysis and development of MRL.
In the light of such a status, we aim to reduce the gap by reviewing existing MRL methods based on their utilized scene map. Specifically, as shown in Fig. \ref{fig:structure}, the MRL methods are categorized and introduced in five classes based on its used map, including not only traditional geo-tagged frame map (Sec. \ref{sec:keyframe}), visual landmark map (Sec. \ref{sec:visual}), and point cloud map (Sec. \ref{sec:geometry}), but also vectorized HD Map (Sec. \ref{sec:hdmap}) and recently raised learnt implicit map (Sec. \ref{sec:nn}). For each kind, we comprehensively review related MRL methods and deeply analyze their strengths, weaknesses, evaluation metrics, and public benchmarks. To further boost the development of the community, we also discuss some opening problems in this field and provide some personal opinions.

{
\vspace{4pt}
\setlength{\parindent}{0cm}
\textbf{Contribution}
The main contribution of this survey can be summarized as follows:
\begin{itemize}
    \item To the best of our knowledge, this manuscript is the first survey exclusively paying attentions on visual localization with monocular camera as main sensor. We attempt to provide an in-depth review so that it can be used as a systematic reference material for newcomers and researchers acquainted with the MRL problem.

    \item The survey reviews MRL methods from a new perspective that we categorize the existing algorithms based on the representation form of utilized map. The relationships between map and MRL methods can be clearly studied.

    \item For each kind of MRL method with diverse scene map, we summarize the typical algorithms but also introduce the popularly used datasets and evaluation metrics.

    \item Additionally, we also provide some opening discussions in MRL researches, which helps the community to further improve the MRL algorithms.

    \item As a final contribution, we build a continuously updated repository at \texttt{\url{github.com}}, including landmark papers reviewed in this survey.

\end{itemize}
}

{
\vspace{4pt}
\setlength{\parindent}{0cm}
\textbf{Organization}
The structure of this survey is as follows. Section \ref{sec:intro} summarizes the contributions and paper structures. Section \ref{sec:background} describes the research background of MRL problem. From Sec. \ref{sec:keyframe} to \ref{sec:nn}, the existing MRL methods are fully reviewed based on their scene map representation. The typical algorithms with various scene map will be evaluated in Sec. \ref{sec:benchmark} and we analyze their advantages and disadvantages, respectively. Finally, Section \ref{sec:discussion} discusses some prominent questions about MRL and section \ref{sec:conclusion} ends with conclusions.
}

\begin{figure*}[!t]
    \centering
    \includegraphics[width=0.97\linewidth]{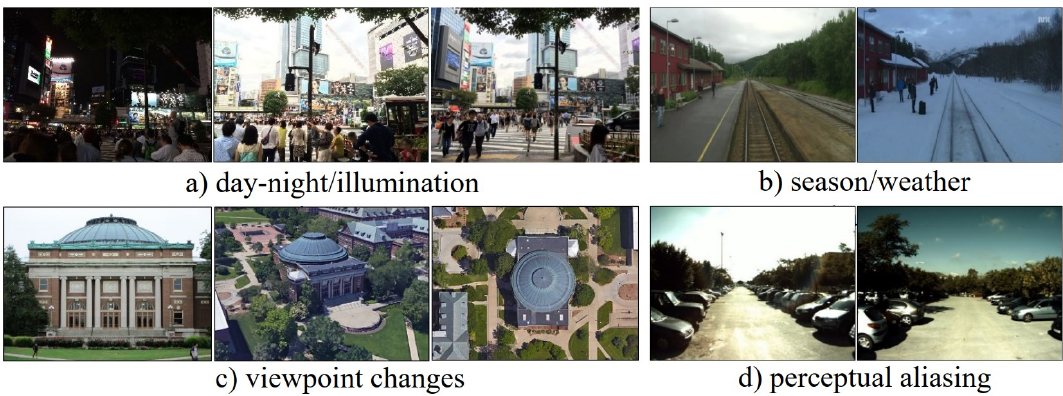}
    \caption{Typical challenges of MRL methods. a) day-night changes with varying illumination \cite{tokyo247}, b) seasonal and weather changes (sunny in summer v.s. snowy in winter) \cite{seqslam}, c) viewpoint changes \cite{university1652}, d) distinct places with visually similar appearances (perceptual aliasing problem) \cite{malaga}.}
    \label{fig:challenges}
\end{figure*}

\section{Background}
\label{sec:background}

\subsection{Problem Formulation and Symbols Definition}
\label{sec:definition}
We start by defining MRL problem in theory and introducing the symbols used in this survey.

During the offline mapping stage, we build the scene map in a specific representation form. In those MRL method that scene map is explicitly represented and saved, scene map can be formatted as $\mathcal{M}_{\mathcal{G}}=\left\{\textbf{m}^{\mathcal{G}}_i\right\}$, where $\textbf{m}^{\mathcal{G}}_i$ is the $i$-th represented item in the map, \textit{e.g.}, visual landmarks, point clouds, and semantics. Here, $\{\cdot\}^{\mathcal{G}}$ denotes the map is considered in a global coordinate $\mathcal{G}$ by default and can be simplified as $\mathcal{M}=\{\textbf{m}_i\}$ if not specified.
When a vehicle $v$ revisits such a mapped area, it captures the observation of the scene, that is, a monocular image $\mathcal{I}$ in MRL research.
The vehicle $v$ tends to estimate its current ego pose based on the current observations $\mathcal{I}$ given scene map $\mathcal{M}$, which is defined as a MRL problem:
\begin{equation}
    \label{eqn:1}
    \hat{\textbf{x}}^{\mathcal{G}}_{v}=\mbox{MRL}(\mathcal{I}~|~\mathcal{M})
\end{equation}
where $\hat{\{\cdot\}}$ denotes an estimated value, $\mbox{MRL}(\cdot|\cdot)$ is a MRL solution, and $\textbf{x}^{\mathcal{G}}_{v}$ is the vehicle's ego pose with regard to $\mathcal{G}$.

Commonly, a 6 DoF pose $\textbf{x}^{\mathcal{G}}_{v}=\left[\textbf{R}^{\mathcal{G}}_{v};~\textbf{t}^{\mathcal{G}}_{v}\right]$ is composed by a 3 DoF rotation $\textbf{R}^{\mathcal{G}}_{v}$ and a 3 DoF translation $\textbf{t}^{\mathcal{G}}_{v}$. The rotation can be represented as quaternion, rotation matrix, or Euler angle. In some MRL application for ground AD vehicles, only 3 DoF pose is considered, \textit{i.e.}, lateral and longitudinal position $[x,y]$ and heading/yaw angle $\theta_{yaw}$.

For better readability, we define frequently used symbols here.
The current image in localization stage is defined as \textit{query} image $\mathcal{I}_q$ while a historical image at $t$ timestamp in mapping stage is defined as \textit{reference} image $\mathcal{I}_{r(t)}$.
Local features from $\mathcal{I}_t$ are denoted as $\mathcal{F}^{t}=\{\textbf{f}^{t}_i\}$ where a local feature $\textbf{f}^{t}_i$ contains a key point $\textbf{p}^{t}_i=[u,v] \in \mathbb{R}^2$ and a corresponding local descriptor $\textbf{d}^t_i \in \mathbb{R}^C$ where $C$ is the dimension.
A pair of matched local features between $\mathcal{I}_q$ and $\mathcal{I}_{r(t)}$ is denoted as $<\textbf{f}^{q}_i,\textbf{f}^{r(t)}_j>$.
The global feature of $\mathcal{I}_t$ is a global descriptor and denoted as $\mathcal{F}_t \in \mathbb{R}^C$.
The items in $\mathcal{M}$ share the same symbols but in global coordinate, \textit{i.e.}, $\{\cdot\}^{\mathcal{G}}$.

\subsection{Challenges of Monocular Re-Localization}
\label{sec:challenges}
MRL has been developed for decades and achieved many great success in many applications. However, tremendous challenges still block the development and usage of MRL methods in real word autonomy. Here, as shown in Fig. \ref{fig:challenges}, some of primary challenges for MRL algorithms are given:

\begin{itemize}
    \item \textbf{Appearance change}: Since monocular camera serves as the primary sensor in MRL solutions, the appearance changes of scene will dramatically affect the algorithms. The conditional appearance variance, \textit{e.g.}, weather \cite{seqslam}, illumination \cite{vprbench,tokyo247}, seasons \cite{seqslam,4seasons}, and day-night changes \cite{benchmark6dof} makes current image visually different from the map, thus challenging matching. And structural changes like dynamic occlusions \cite{slcd}, and layout changes \cite{InLoc} will interfere with the geometrical pose estimation.

    \item \textbf{Viewpoint difference}: When the camera viewpoint is quite different between current timestamp and mapping stage, MRL method will be struggled by limited co-visible area and changed layout of items. As typical instances, cross-view MRL for drones \cite{each_part_matters,university1652}, opposite-view \cite{donot_look_back,garg2018lost} and even ground-to-aerial \cite{xview} MRL for ground vehicles attract growing interests.
    
    \item \textbf{Perceptual aliasing}: In some scenarios with visually similar or repeated textures, the MRL method will generate ambiguous estimation when distinct places have similar appearances, \textit{e.g.}, corridors \cite{rapnet,openloris}, and parking lots \cite{malaga}, which is called perceptual aliasing problem.

    \item \textbf{Generalization and scalability}: The real world scenarios are infinite, and we cannot exhaust all the types, visual conditions, and interference of possibly occurred scenes. A practical real-world autonomy requires MRL methods to work stably in diverse scenarios, even in unseen environments, called generalization ability \cite{keetha2023anyloc,SANet}. Besides, the real world is unbounded, we also need MRL solutions scalable, which limits the unbearable increasement of map size and computational costs with vehicles continuously explore.
\end{itemize}

\subsection{Comparison with other surveys}
\label{sec:comparision with other survey}

In decades of years, numerous reviews on autonomous robotics techniques have been raised, addressing various topics, \textit{e.g.}, place recognition \cite{LCD2022TITS,Yin2022GeneralPR,VPR_a_survey,vpr_survey_ijcai,VRP_a_survey_from_DL_perspective}, ego and object localization \cite{VBL2018PR,a_survey_on_visual_map_localization_using_LiDARS_and_camera,visual_and_object_geo_localization_a_comprehensive_survey}, SLAM \cite{past_present_future_SLAM,a_survey_on_DL_for_SLAM,visual_SLAM}.
As a milestone survey, Lowry \textit{et al.} \cite{VPR_a_survey} firstly introduced the concepts behind visual place recognition. Later, Garg \textit{et al.} \cite{vpr_survey_ijcai} refined the definition by introducing orientation limitation. With a growing number of visual place recognition methods based on Deep Learning (DL), Zhang \textit{et al.} \cite{VRP_a_survey_from_DL_perspective} reviewed recently proposed solutions from the DL perspective. Yin \textit{et al.} \cite{Yin2022GeneralPR} presented general place recognition challenges. However, place recognition methods could only serve as a coarse localization that provide a rough pose approximation, which only cover few parts of visual localization researches (Sec. \ref{sec:VPR} in this survey). Place recognition methods are reviewed and called indirect MRL methods in \cite{VBL2018PR}, and the authors also provide direct MRL methods that give the exact 6 DoF pose of the \textit{query} according to a known \textit{reference}, which corresponds to Sec. \ref{sec:visual}, \ref{sec:APE}, and \ref{sec:SCR} in this survey.
In \cite{a_survey_on_visual_map_localization_using_LiDARS_and_camera}, Elhousni \textit{et al.} defined visual map localization as a two-stage process, namely, place recognition and metric map localization, and reviewed MRL methods using LiDAR, camera, or cross-modal sensors. Such a two-stage framework is only adopted by those MRL solutions using visual landmark map and point cloud map, and it cannot completely describe all the MRL solutions, such as newly emerged absolute pose regression-based methods \cite{posenet}, scene coordinate regression-based methods \cite{neumap}, \textit{etc.} 

By deeply analyze existing reviews about MRL methods, we find that although all these surveys have made great contributions to the community and boosted the progress of MRL research, they have regrettably neglected the relationship between MRL methods and scene map, and thus cannot analyze MRL solutions in a unified framework. 
In this paper, we propose to review MRL methods from the perspective of scene map so that we can clearly categorize them based on the representation form of utilized map, thereby providing a comprehensive and deep review about MRL methods.

\section{Geo-tagged Frame Map}
\label{sec:keyframe}

\begin{figure}[!t]
    \centering
    \includegraphics[width=0.97\linewidth]{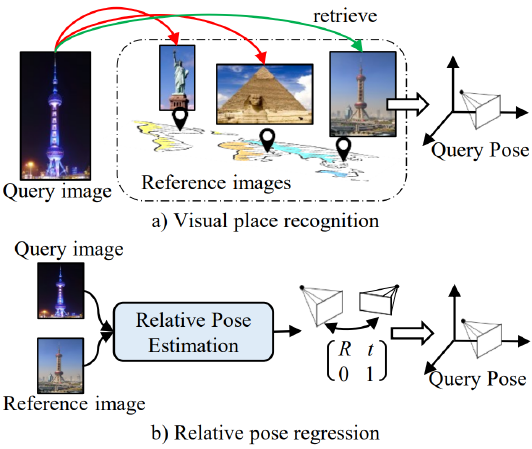}
    \caption{A diagram of MRL methods using geo-tagged frames as map, including a) VPR and b) RPR methods.}
    \label{fig:vpr_rpr}
\end{figure}

During mapping stage, many frames are labeled by geographic position or precise 6 DoF poses. These geo-tagged frames can be served as scene map in MRL methods. In this field, some proposals estimate ego pose by historical pose of retrieved geo-tagged frames as shown in Fig. \ref{fig:vpr_rpr} a) and others estimate relative pose between \textit{query} image and geo-tagged frames as shown in Fig. \ref{fig:vpr_rpr} b).

\subsection{Visual Place Recognition}
\label{sec:VPR}

Visual Place Recognition (VPR) algorithm aims to identify the re-observed places by retrieving \textit{reference} frame when the vehicle goes back to a previously visited scene, which is also adopted for loop closure detection in SLAM system \cite{LCD2022TITS}. 
After obtaining best-matched \textit{reference} image, VPR regards the pose of retrieved \textit{reference} frame as an approximated pose of current \textit{query} image. 

Formally, in VPR, the scene map $\mathcal{M}$ is represented by \textit{reference} frames $\mathcal{I}_{r(t)}$ with pose $\textbf{x}^{\mathcal{G}}_t$, \textit{i.e.}, $\mathcal{M}=\{\textbf{m}_t\}$ where $\textbf{m}_t=<\mathcal{I}_{r(t)},\textbf{x}^{\mathcal{G}}_t>$.
Given \textit{query} image $\mathcal{I}_q$, a VPR algorithm 
\begin{itemize}
    \item[1] represent $\mathcal{I}_q$ and all the $\mathcal{I}_{r(t)}$ in $\mathcal{M}$ as features:
        \begin{equation}
            \begin{split}
                \label{equ:vpr1}
                \mathcal{F}^{q}&=\mbox{ImFeat}\left(\mathcal{I}_q\right)~~\mbox{for \textit{query}} \\
                \mathcal{F}^{r(t)}&=\mbox{ImFeat}\left(\mathcal{I}_{r(t)}\right)~~\mbox{for \textit{reference}}
            \end{split}
        \end{equation}
    \item[2] retrieve the best-matched \textit{reference} image based on the similarity of features:
        \begin{equation}
            \label{equ:vpr12}
            \hat{\textbf{m}}=\mathop{\arg\max}\limits_{\mathbf{m}_i} P(\{\mathcal{F}^{q}, \mathcal{F}^{i}~\mbox{of}~\textbf{m}_i | \textbf{m}_i \in \mathcal{M})
        \end{equation}
\end{itemize}
where the matching function $P(A,B)$ provides the matching similarity between $A$ and $B$, and $\textbf{m}_i$ is a matching candidate, $\hat{\textbf{m}}$ is the best-matched one for $\mathcal{I}_q$. Then, the pose of current \textit{query} image is given by the pose of retrieved \textit{reference} frame:
\begin{equation}
    \label{equ:vpr2}
    \hat{\textbf{x}}^{\mathcal{G}}_q \leftarrow \textbf{x}^{\mathcal{G}}_{\hat{\textbf{m}}}
\end{equation}

In early proposals, SeqSLAM \cite{seqslam} and Fast-SeqSLAM \cite{fastslam} directly use down-sampled and normalized image patches to measure image similarity. However, raw image intensity is sensitive to visual interference. So VPR methods often extract high-level image features from images and measure feature similarity as a more reliable image similarity. In this feature-based scheme, image feature algorithm plays an important role in VPR solutions. In VPR, based on the receptive field, the image feature can be categorized into global feature \cite{hog,vlad,netvlad} and local feature \cite{sift,superpoint}. 

\begin{figure*}[!h]
    \centering
    \includegraphics[width=0.97\linewidth]{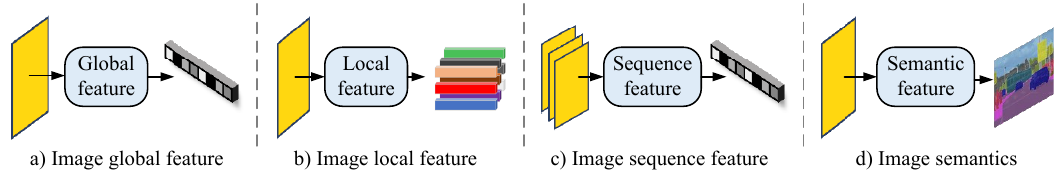}
    \caption{Four kinds of image features in VPR, including a) image global feature, b) image global feature, c) image sequence feature, and d) image semantics.}
    \label{fig:vpr_feature}
\end{figure*}

{
\vspace{4pt}
\setlength{\parindent}{0cm}
\textbf{Global Feature-based VPR (GF-VPR):}
\label{sec:global feature VPR}
Global feature algorithm directly represent the whole image as a compact matrix or vector so the image similarity can be simply measured as a cosine distance 
\begin{equation}
P(A,B)=({A\cdot B})/({\Vert{A}\Vert \Vert{B}\Vert})
\end{equation}
Therefore, the core component of the GF-VPR methods is the feature algorithm $\mbox{ImFeat}(\cdot)$. 
Before the DL stage, the global image features utilized in this area is designed to analyze the statistical distribution of image intensity, \textit{e.g.}, Gist \cite{Siagian}, color histograms \cite{Ulrich}, VLAD \cite{vlad}, BoW \cite{bow}, ASMK \cite{ASMK}, HOG \cite{hog}, and improved CoHOG \cite{CoHOG}.
As a widely used instance, VLAD \cite{vlad} aggregates $N$ local descriptors $\{\textbf{d}_i\}^{N}_{i=1}$ into a compact matrix $V$ using a codebook with $K$ clusters:
}
\begin{equation}
    \label{equ:vlad}
    V(j,k)=\sum^{N}_{i=1} a_k(\textbf{d}_i)\left(\textbf{d}_i(j)-\textbf{c}_k(j)\right)
\end{equation}
where $\textbf{d}_i(j)$ and $\textbf{c}_k(j)$ are the $j$-th dimensions of the $i$-th descriptor and $k$-th cluster center, respectively. $a_k(\textbf{d}_i)$ denotes the membership of the $\textbf{d}_i$ to $\textbf{c}_k$, \textit{i.e.}, $a_k(\textbf{d}_i)=1$ if $\textbf{c}_k$ is the closest cluster to $\textbf{d}_i$ and 0 otherwise.

Since DL-based methods dominate the visual tasks, some VPR methods have applied Deep Neural Network (DNN) to extract global features. As a milestone of DNN-based global feature, NetVLAD \cite{netvlad} designs a trainable feature aggregation layer based on the principle of VLAD algorithm \cite{vlad}.
It replaces the non-differentiable function $a_k(\textbf{d}_i)$ with soft-assignment of descriptors to multiple clusters, and thus construct an fully-differentiable NetVLAD layer:
\begin{equation}
    \label{equ:netvlad}
    V(j,k)=\sum^{N}_{i=1} \frac{e^{w^T_k \textbf{d}_i+b_k}}{\sum_{k'} e^{w^T_{k'} \textbf{d}_i+b_{k'}}}\left(\textbf{d}_i(j)-\textbf{c}_k(j)\right)
\end{equation}
where $\{w_k\},\{b_k\}$, and $\{\textbf{c}_k\}$ are trainable for each cluster $\textbf{c}_k$.

Meanwhile, many other attempts are made by scholars to boost the representative ability of global feature. GeM \cite{GeM} designs a trainable generalized-mean pooling layer that generalizes maximum and average pooling for feature aggregation. TransVPR \cite{transvpr} could aggregate task-relevant features by self-attention mechanism in Transformers \cite{transformer}. MixVPR \cite{mixvpr} incorporates a global relationship between elements in each feature map in cascaded feature mixing modules. AnyLoc \cite{keetha2023anyloc} leverages an off-the-shelf self-supervised visual foundation model features (such as DINOv2 \cite{oquab2023dinov2}) with no VPR-specific training or finetuning. 

Although global feature can compactly represent images, the retrieval time of GF-VPR methods will linearly increase with the size of the database. And the robustness against viewpoint changes and dynamic occlusions of global features is also inferior than local feature \cite{DSFeat}.

{
\vspace{4pt}
\setlength{\parindent}{0cm}
\textbf{Local Feature-based VPR (LF-VPR):}
\label{sec:local feature VPR}
A local feature algorithm extracts massive key points and corresponding descriptors from images. Then, the extracted local features are utilized to retrieve the best-matched \textit{reference} frames. In early LF-VPR researches, $\mbox{ImFeat}(\cdot)$ is traditional hand-crafted local features \cite{sift,orb,surf}. Then, DNN-based local features are adopted in VPR tasks. Early researches directly apply off-the-shelf DNN model pretrained by other computer vision tasks as local feature extraction module \cite{BoCNF,PlaceRW,convnet_pr}, and recent methods leverage localization task-specific local features and achieve improved performances \cite{superpoint,fild2}. In dynamic scene, local features located at dynamic objects/regions will confuse VPR methods, so Chen \textit{et al.} \cite{slcd} removed inconsistent dynamics by exploiting instance-level semantics. But semantics are hard to accurately judge the actual motion state of the objects, so DSFeat \cite{DSFeat} learns to select stable and discriminative local features based on not only semantics but also attentions. 
}

Directly matching between local features and using overall matching similarity is a straightforward way to measure image similarity, but it is computational costly. Local features need to be further processed to measure image similarity, so the design of $P(A,B)$ is the core component of LF-VPR methods.
One common strategy is to aggregate extract local features to a compact global feature, \textit{e.g.}, VLAD \cite{vlad,multivlad}, ASMK \cite{ASMK}, and BoW \cite{bow}, and performance retrieval as GF-VPR methods.
As typical solutions. DBoW \cite{dbow-iros,dbow-tro} builds tree-shaped vocabulary to aggregate local features so that an image can be compactly represented by the occurrences of visual words (some general local features) in the vocabulary, following the principle of BoW \cite{bow}. And in \cite{LRO}, authors adopted ASMK-based framework\cite{ASMK} to aggregate local features and retrieve candidates. Compared with the BoW \cite{bow}, ASMK \cite{ASMK} can eliminate the problem of quantization noise and achieve improved precision.
In addition to those LF-VPR solution aggregating local features and retrieving like GF-VPR methods, some other methods adopt probabilistic scheme to find matched \textit{reference} frame in $P(A,B)$. FAB-MAP \cite{fabmap} and FAB-MAP 2.0 \cite{fabmap2} train a Chow-Liu tree to learn the co-visibility of features and then estimate the matching probability of candidates.
IBuILD \cite{ibuild} tracks local features along video streams to incrementally generate vocabulary, and perform VPR by a likelihood function. 
iBoW-LCD \cite{ibow-lcd} applies Bayesian framework by measuring the importance of local features with regard to the vocabulary and \textit{query} image. Then, LiPo-LCD \cite{lipolcd} improves by adding line-wise local feature in human-made environments.
Tsintotas \textit{et al.} found \textit{reference} images by a binomial probability density function where local features are assigned to images as votes \cite{AVWs,BoTW}.

Merely using descriptor information of local features will confuse distinct places with visually similar textures, called perceptual aliasing problem as show in Fig. \ref{fig:challenges} d). So, many advanced methods exploit geometrical information from key points of local features as a re-ranked matching score or post verification after obtaining candidate \textit{reference} images by Eqn. \ref{equ:vpr12}.
Patch-NetVLAD \cite{patch-netvlad} matches patch-level local DNN features as geometrical verification to re-order candidates retrieved by NetVLAD \cite{netvlad}. 
As a de-facto standard in LF-VPR, researchers apply epipolar check across images and validate the matched \textit{reference} by the number of feature inliers \cite{dbow-iros, dbow-tro, htmap, AVWs, BoTW, fild2}. This method works properly in most scenes but fails if mismatched local features were dominated or pre-defined parametric model was incorrect.
So, topological graph-based post-verification \cite{yue1,yue2} and local relative orientation matching \textit{et al.} \cite{LRO} are developed to measure geometrical consistency.
Some other works introduce temporal consistency as post-verification \cite{ibuild, ibow-lcd, AVWs}, but this scheme is limited in image sequence. They follow a basic assumption that if $I_{q(t_i)}$ and $I_{r(t_j)}$ were matched, their adjacent images $I_{q(t_i-1)}$ and $I_{r(t_j-1)}$ should also be matched.

\begin{figure*}[!t]
    \centering
    \includegraphics[width=0.97\linewidth]{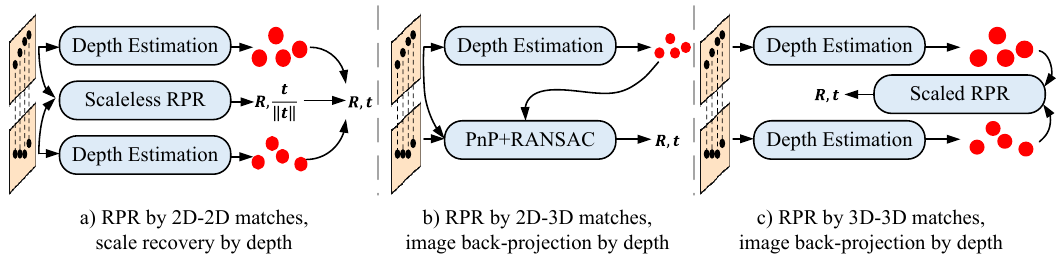}
    \caption{Three G-RPR solution proposed in \cite{arnold2022mapfree}, namely, a) first solve scaleless relative pose by 2D-2D matches then recover scale by 3D-3D matches using estimated depth, b) first lift 2D-2D matches to 2D-3D matches by depth then estimate scaled pose by PnP+RANSAC solver, c) lift 2D-2D matches to 3D-3D matches using depth then estimate scaled pose.}
    \label{fig:rpr}
\end{figure*}

{
\vspace{4pt}
\setlength{\parindent}{0cm}
\textbf{Sequence Feature-based VPR (SF-VPR):}
\label{sec:sequence feature VPR}
SeqNet \cite{seqnet} and Delta Descriptor \cite{delta-des} proposes to measure the changes across image global features as sequence global feature. SeqVLAD \cite{SeqVLAD} combines NetVLAD layer \cite{netvlad} and sequence global features. In these works, VPR first performs sequential matching across the image sequences then finds the most matched frame within the matched sequence. The SF-VPR is technically similar to GF-VPR but applied in image sequences.
}

{
\vspace{4pt}
\setlength{\parindent}{0cm}
\textbf{Semantic-based VPR (S-VPR):}
\label{sec:semantic VPR}
Gawel \textit{et al.} \cite{xview} utilized semantic objects as nodes, and built a semantic topological graph based on the nearby relationship of semantic objects in images. The graph can be compactly represented by a global feature based on random walk \cite{deepwalk}. Also, S-VPR is performed as GF-VPR but global feature is calculated by semantics.
}

{
\vspace{4pt}
\setlength{\parindent}{0cm}
\textbf{Summary:}
\label{sec:VPR summary}
VPR algorithms retrieve geographical nearby \textit{reference} frame to localize. Benefit from robustly representing images and map, current VPR solution can fully and precisely retrieve the frame. However, VPR achieves localization by regarding the \textit{reference} pose as an approximated \textit{query} pose, which inherently limits the localization accuracy when the current trajectory is very different from the mapping trajectory. Thus, VPR usually serves as a coarse or initial localization step in high-precision localization system \cite{hloc1,hloc2,LATITUDE}.
}

\subsection{Relative Pose Estimation}
\label{sec:RPR}

Relative Pose Estimation (RPR) methods aims to estimate the relative pose between \textit{query} image and \textit{reference} image in the map. In this field, the scene map is also represented by images with pose, \textit{i.e.}, $M=\{\textbf{m}_t\}$, where $\textbf{m}_t=<\mathcal{I}_{r(t)},\textbf{x}^\mathcal{G}_t>$. The scene sometimes only contain one \textit{reference} image $M=\textbf{m}_r=<\mathcal{I}_r,\textbf{x}^\mathcal{G}_r>$ \cite{arnold2022mapfree}. When the \textit{query} image $\mathcal{I}_q$ is fed into the system, the RPR algorithm estimate the relative pose of $\mathcal{I}_q$ with regard to its (closest) \textit{reference} image $\mathcal{I}_r$:
\begin{equation}
    \label{equ:rpr1}
    \hat{\textbf{x}}^r_q = \mbox{RPR}(\mathcal{I}_q, \mathcal{I}_r)
\end{equation}
so that the absolute pose of $\mathcal{I}_q$ can be calculated by:
\begin{equation}
    \label{equ:rpr2}
    \hat{\textbf{x}}^\mathcal{G}_q \leftarrow \textbf{x}^\mathcal{G}_r * \hat{\textbf{x}}^r_q
\end{equation}
where $*$ is the pose transform operator. 
RPR methods avoid costly pre-scanning and reconstructing scene, enabling economical MRL in some new environment with extremely sparse records, which is commonly seen in Augmented Reality (AR) applications: User A shares a photo with its location in a new scene, any user B can instantly re-localize with regard to user A and perform interactions \cite{arnold2022mapfree}. Generally, RPR methods can be categorized into geometry-based methods and learning-based methods. 

{
\vspace{4pt}
\setlength{\parindent}{0cm}
\textbf{Geometry-based RPR (G-RPR):}
\label{sec:geometry-based RPR}
The G-RPR methods perform feature matching between \textit{query} image and \textit{reference} image to obtain pixel-level correspondence, and then heuristically solve relative pose. 
According to Multiple View Geometry (MVG) theory \cite{MVG}, the relationship between the 2D projection of a 3D point on camera coordinates located on two views can be formatted as an essential matrix $E$:
\begin{equation}
    \label{equ:ematrix1}
    {(K_1\textbf{p}_1)}^{T} E (K_2 \textbf{p}_2) = 0
\end{equation}
where $\textbf{p}_1$ and $\textbf{p}_2$ denotes 2D projection in image coordinate (key point). And the essential matrix actually contains the relative rotation and translation between two views:
\begin{equation}
    \label{equ:ematrix2}
    E = \textbf{t} \times \textbf{R}
\end{equation}
So, G-RPR problem with known camera intrinsic can be solved by first matching local features, then estimating an essential matrix, and finally decomposing the essential matrix to the relative pose (including relative rotation and scaleless translation) between two images. Later proposals further improve such a basic formula by applying better features \cite{sift,surf,orb,superpoint,disk,alike,aliked,r2d2}, better matching \cite{superglue,adalam,lindenberger2023lightglue} or better robust estimators \cite{ransac_analysis,good_correspondence,MAGSAC,MAGSAC++,NGRANSAC,ACNe}.
}

Since the relative pose decomposed from Essential matrix is up-to-scale, so many scholars utilized scaleless pairwise relative poses between the \textit{query} image and multiple \textit{reference} images to triangulate the scaled, metric pose of \textit{query} image in the early G-RPR solutions \cite{EssNet,ExReNet}. 
However, in the scene where only one \textit{reference} image exists, such a scheme fails to recover scale. 
In \cite{arnold2022mapfree}, as show in Fig. \ref{fig:rpr}, authors proposed to utilized estimated monocular depth \cite{DPT,planeRCNN} to a) provide scale or b) lift 2D-2D correspondences to 2D-3D or c) 3D-3D correspondences, and concluded the second solution back-projecting \textit{reference} image to 3D space and estimating scaled relative pose by 2D-3D matches achieves best performance, which is technically converting G-RPR to typical Perspective-n-Point (PnP) problem and quite similar to MRL with visual landmark map that we will discuss later in Sec. \ref{sec:visual}. And \cite{arnold2022mapfree} also experimentally concludes better feature matching could boost the G-RPR methods.

{
\vspace{4pt}
\setlength{\parindent}{0cm}
\textbf{Learning-based RPR (L-RPR):}
\label{sec:learning-based RPR}
Using DNN model, L-RPR algorithms can predict the relative pose from two input images in an E2E way, eliminating the requirement of explicit local feature matching and pose recovery. Such a scheme decreases the requirement of scene texture and co-visible area between images, so it can work well in some challenging scenarios that G-RPR approaches fails due to disabled module. For example, in \cite{Cai2021ExtremeRot, DirectionNet}, authors focused on develop RPR in extreme cases, including when there is low or even no overlap between the query image and reference image. Additionally, DNNs empower L-RPR methods the ability of scale recovery. Despite some methods still estimate pose up-to-scale and recover scales by triangulation \cite{CNNSPP,ExReNet,Zakaria2017ICCVW}, there are some proposals directly estimating metric relative pose between \textit{query} image and \textit{reference} image \cite{DistillPose,RelocNet,RPNet}. These two strategies can both show pleased generalization ability on unseen scenarios.
}

In \cite{arnold2022mapfree}, authors combined G-RPR and L-RPR strategies, and designed a network to predict 3D-3D correspondence between two images. The relative pose is then solved as G-RPR method using Orthogonal Procrustes \cite{Procrustes} inside a RANdom SAmple Consensus (RANSAC) loop \cite{ransac}. The experimental results claimed that this solution outperforms the L-RPR method, indicating that involving MVG knowledge in DNN may benefit RPR algorithms. 

{
\vspace{4pt}
\setlength{\parindent}{0cm}
\textbf{Summary:}
\label{sec:RPR summary}
RPR perform localization by estimating relative pose between \textit{query} and \textit{reference} images. Since scale recovery for pairwise RPR is not perfectly solved yet, RPR methods cannot achieve very high-precision localization performance. But, the low requirement of scene map, and slightly higher localization accuracy than VPR still let RPR attract growing attention in light applications like AR.
}

\begin{figure}[!h]
    \centering
    \includegraphics[width=0.97\linewidth]{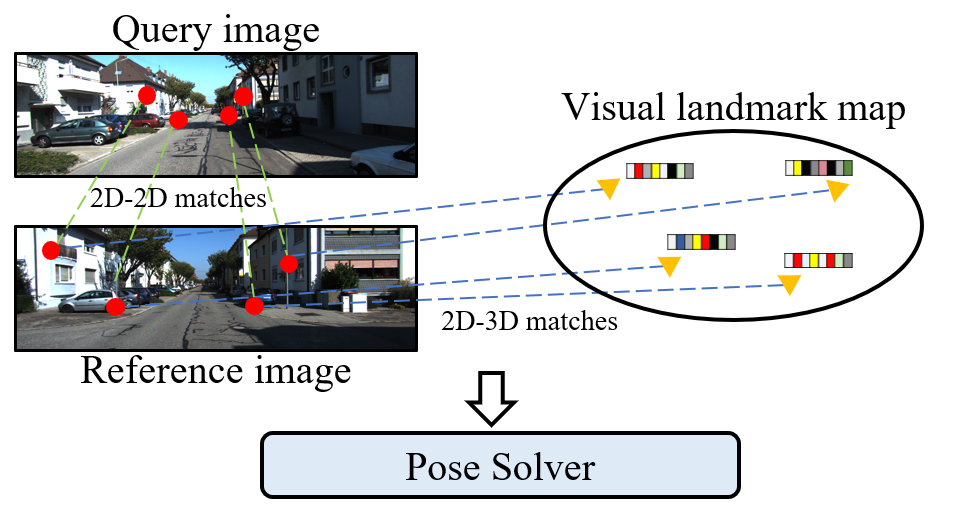}
    \caption{A diagram of MRL method with visual landmark map.}
    \label{fig:visual}
\end{figure}

\section{Visual Landmark Map}
\label{sec:visual}

As the most popularly utilized representation format of scene map, visual landmark map is generally constructed by Structure from Motion (SfM) \cite{colmap} or SLAM \cite{orbslam3}. Here, visual landmarks are some informative and representative 3D points that lifted from 2D pixels by 3D reconstruction, and they are associated with corresponding local features in various observed images as shown in Fig. \ref{fig:visual}. Formally, scene map can represented as $\mathcal{M}=\{\textbf{m}_i\}$, where $\textbf{m}_i=<\textbf{P}^{\mathcal{G}}_i,\{\mathcal{F}^{t}_{i}\}_t>$ is a visual landmark, $\textbf{P}^{\mathcal{G}}_i$ denotes the 3D location of $\textbf{m}_i$ in global frame $\mathcal{G}$, $\mathcal{F}^{t}_{i}$ is a set of local features from \textit{reference} image $\mathcal{I}_{r(t)}$ associated with $\textbf{m}_i$. Among the Visual Landmarks-based MRL methods (VL-MRL), Hierarchical Localization (HLoc) \cite{hloc1,hloc2} is the most famous framework that has been widely used in many applications. Up to now, it still dominates the long-term visual localization task\footnote{https://www.visuallocalization.net/benchmark/}.

In HLoc framework \cite{hloc1,hloc2}, visual landmark map is offline built via SfM reconstruction \cite{colmap}. During online localization stage, \textit{query} image $\mathcal{I}_q$ is matched with retrieved \textit{reference} image $\mathcal{I}_r$ and the resulting 2D-2D matches are lifted to 2D-3D matches between $\mathcal{I}_q$ and visual landmarks based on $\mathcal{M}$, which can be used to solve scaled pose as a typical PnP problem. For clarify, we split VL-MRL algorithms into two-step pipeline (Step 1.A and 1.B are alternative):

\begin{itemize}
    \item[1.A] extract local feature from $\mathcal{I}_q$ and $\mathcal{I}_r$ (we assume only one $\mathcal{I}_r$ is utilized here):
        \begin{equation}
            \begin{split}
                \label{equ:hloc1}
                \mathcal{F}^{q} = \{\textbf{f}^{q}_i\}_i &=\mbox{ImFeat}(\mathcal{I}_q)~~\mbox{during localization} \\
                \mathcal{F}^{r} = \{\textbf{f}^{r}_i\}_i &=\mbox{ImFeat}(\mathcal{I}_r)~~\mbox{during mapping}
            \end{split}
        \end{equation}
        Then, match local features between $\mathcal{I}_q$ and $\mathcal{I}_r$ and obtain 2D-2D correspondences:
        \begin{equation}
            \label{equ:hloc2}
            \{<\textbf{f}^{q}_{i},\textbf{f}^{r}_{j}>\} = \mbox{Match} (\mathcal{F}^{q},\mathcal{F}^{r})
        \end{equation}
        where $i,j$ are the indices.
    \item[1.B] or jointly extract and match local feature between images:
        \begin{equation}
            \label{equ:hloc3}
            \{<\textbf{f}^{q}_{i},\textbf{f}^{r}_{j}>\} = \mbox{FM} (\mathcal{I}_q, \mathcal{I}_r)
        \end{equation}
    \item[2] get the associated visual landmarks of $\mathcal{F}^{r}$ in $\mathcal{M}$ so that 2D-2D correspondences can be lifted to 2D-3D correspondences, and the pose can be solved as a typical PnP problem:
        \begin{equation}
            \begin{split}
                \label{equ:hloc4}
                \hat{\textbf{x}}^{\mathcal{G}}_{q} &= \mbox{SolvePnP} (\{<\textbf{p}^{q}_{i},\textbf{p}^{r}_{j}>\}~|~\mathcal{M}) \\
                &=\mbox{SolvePnP} (\{<\textbf{p}^{q}_{i},\textbf{P}^{\mathcal{G}}_{j}>\})
            \end{split}
        \end{equation}
\end{itemize}

In this section, we review related algorithms in these two steps of VL-MRL methods, namely, local feature extraction then matching (Step 1.A), joint local feature extraction and matching (Step 1.B), and pose solver (Step 2). And we also review some proposals aiming at other perspectives in VL-MRL methods.

\subsection{Local Feature Extraction-then-Matching}
\label{sec:local feature extraction then matching}

{
\vspace{4pt}
\setlength{\parindent}{0cm}
\textbf{Local Feature Extraction:}
\label{sec:local feature extraction}
Local feature detects massive salient pixels in an image (denoted as key points), and describes the neighboring area of the key point using a high-dimensional vector (denoted as descriptor). 
Under the context of VL-MRL methods, local feature is utilized to obtain accurate 2D-2D pixel-wise correspondences between images. The key points detected by local feature algorithms should be repeatable under different image conditions and qualities while located in salient regions \cite{r2d2} so that visual landmarks built by local features can be sparse but informative to represent the whole scene \cite{rapnet,DSFeat}. And descriptors should be designated to enable accurate correspondence establishment between detected local feature across different views \cite{l2net,hardnet,sosnet,geodesc,contextdesc}, so they need be invariant against typical visual interference. Additionally, the positions of local features should be accurate \cite{aslfeat,alike,aliked}. 
}

The early local features are built upon the human prior knowledge about the quality of pixels. As milestones in this area, float-valued SIFT \cite{sift}, SURF \cite{surf} and binary-valued BRISK \cite{brisk}, BRIEF \cite{brief}, ORB \cite{orb} have been widely used and achieved great successes. However, due to the limited presentation ability, these hand-crafted local features are challenged by complex visual interference in real-world scenarios. 
Since entering the perception age ruled by DNN, researchers have begun to introduce DL technique in local feature extraction. 
As a milestone, SuperPoint \cite{superpoint} was proposed in 2018 and it leverage a Convolutional Neural Network (CNN) model to extract local features in an E2E way. The utilized CNN model is composed of a shared encoder and two heads. The encoder extracts high-dimensional dense features, and one head generate pixel-wise score map to detect key points while the other head generate descriptors. Such a network architecture is widely used in later works \cite{r2d2,sekd,alike,aliked,disk} and called ``detect-then-describe'' framework. D2-Net \cite{d2net} and ASLFeat \cite{aslfeat} adopt a different CNN architecture that only a encoder is used to generate dense descriptor map, and the score map used to detect key point is calculated from the descriptor map, called ``detect-and-describe'' framework.

The most important part in designing DL-based local feature is its training scheme, which determines the performance of local features, \textit{e.g.}, repeatability and accuracy of key points, robustness and discriminativeness of descriptors. SuperPoint \cite{superpoint}, as a ganger, is first trained on synthetic samples and then fine-tuned in two images generated by its proposed homographic adaptation techinique. Later, some real-world datasets \cite{megadepth} with posed images and 3D map are proposed, the ground-truth correspondences between different images can be available, facilitating the training of DL-based local features. Most algorithms apply metric learning between two co-visible patches or images to optimize their network parameters so that the network can gradually learn to extract local features suitable for feature matching \cite{d2net,aslfeat,alike,aliked,r2d2}. So, their key points are repeatable while descriptors can be precisely matched. To make DNN-based local feature get rid of costly training data collection, SEKD \cite{sekd} proposes a self-evolving framework to train CNN model without any annotations or pre-processing on the training data. DISK \cite{disk} trains CNN model not by metric learning but by reinforcement learning scheme where the feature extraction and matching procedures can be trained in an E2E manner, thereby greatly boosting the performance of local features. To detect informative key points in images, some methods regard the score map as attention mask to aggregate dense descriptor, so the local feature algorithms can be trained as GF-VPR methods, which also brings significant improvements \cite{delf,mda,DSFeat}.  

\begin{figure}[!b]
    \centering
    \includegraphics[width=0.97\linewidth]{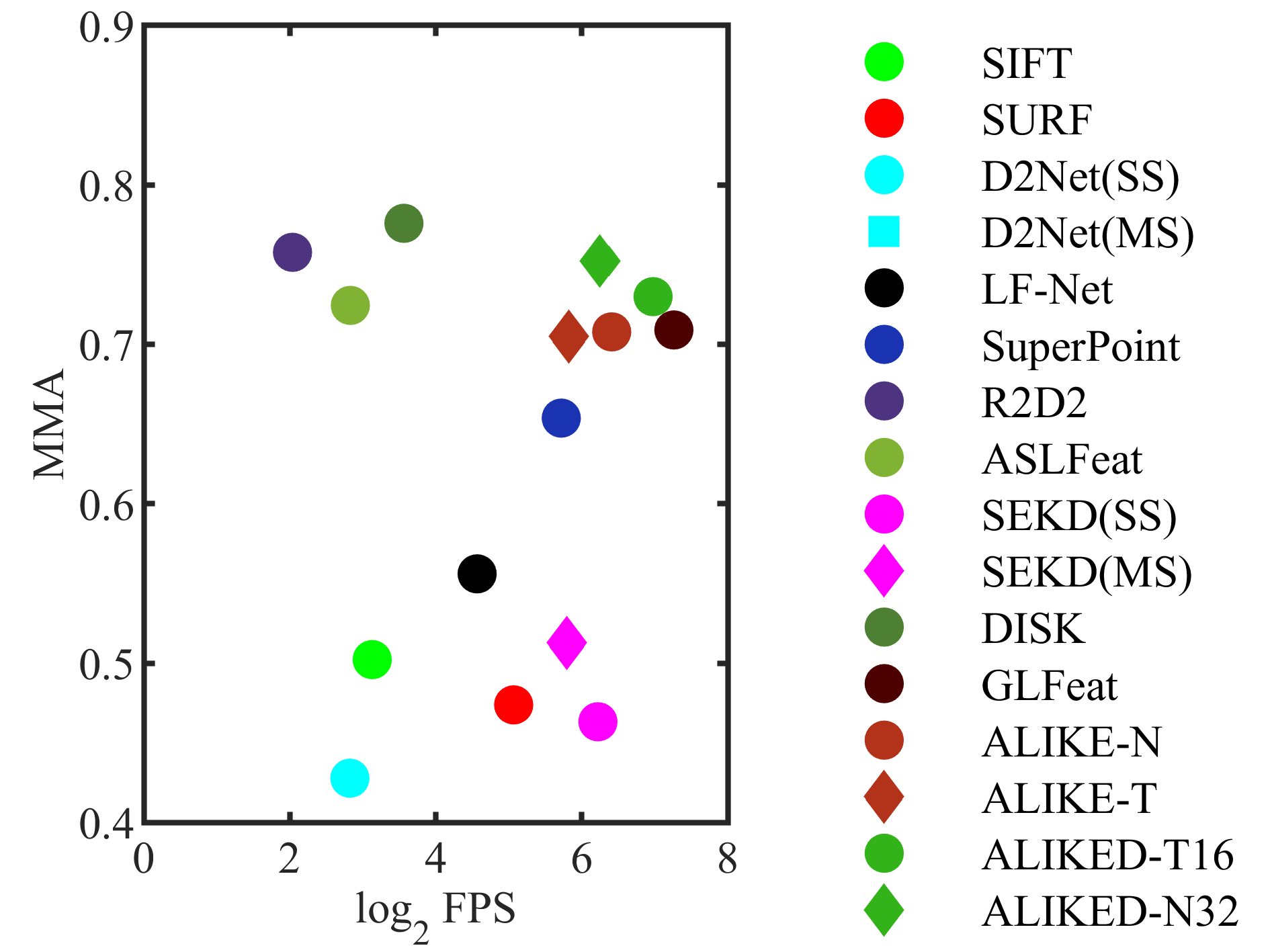}
    \caption{The processed Frame Per Second (FPS) and Mean Matching Accuracy (MMA) on HPatches dataset \cite{hpatches1,hpatches2}. The FPS metrics of SIFT \cite{sift}, SURF \cite{surf}, D2-Net \cite{d2net}, SEKD \cite{sekd}, GLFeat \cite{glfeat} are evaluated on a NVIDIA RTX GPU \cite{glfeat}, whereas LF-Net \cite{lfnet}, SuperPoint \cite{superpoint}, R2D2 \cite{r2d2}, ASLFeat \cite{aslfeat}, DISK \cite{disk}, ALIKE \cite{alike}, and ALIKED \cite{aliked} are evaluated on a NVIDIA GeForce RTX 2060 GPU \cite{aliked}. ``SS'' denoted single scale detection while ``MS'' is multiple scale detection.}
    \label{fig:time-mma}
\end{figure}

Limited by the inherent locality of CNN, most existing local features description methods only learn local descriptors with local information and lack awareness of global context. Thus, MTLDesc \cite{MTLDesc} and GLFeat \cite{glfeat} aggregate non-local information into local feature extraction. Besides, as shown in Fig. \ref{fig:time-mma}, most of the State-of-the-Art (SOTA) local features cannot meet the requirements of efficiency even with high-performance GPU devices, which blocks the usage of local feature algorithms in real-time applications. Zhao \textit{et al.} specially designed light CNN model for real-time local features \cite{alike,aliked}. Thin neural networks usually sacrifice representative ability, so strong supervision for local feature training is required to achieve balance between matching accuracy and computational efficiency \cite{alike,aliked,glfeat}.

Monocular images lack of actual scale information, which affect the performance of local features. As a commonly applied strategy, Multiple Scale (MS) detection can help local feature extraction. Some works explicitly apply MS detection \cite{d2net,sekd}, while some other works implicitly apply MS detection in network by fusing multiple features from feature pyramid inherited by CNN model \cite{aslfeat,alike}. 

{
\vspace{4pt}
\setlength{\parindent}{0cm}
\textbf{Local Feature Matching:}
\label{sec:local feature matching}
Before DL stage, researchers often apply the brute-force Nearest Neighbour (NN) searching to find matches between the extracted local features and perform ratio test to reject mismatches \cite{sift}. Then, FLANN algorithm \cite{FLANN} is developed as an efficient alternate of brute-force searching. To get more robust matches, some methods explore positional distribution of local features \cite{LPM,Ma2018GLP,CODE}, and some other works further explore locally adaptive local descriptor selection \cite{Hu2015TIP} and optical flow guided matching \cite{Maier2016GuidedMB} for better matching. These methods only consider local properties of local features so they can hardly achieve globally consistent matching. So, some approaches turn to leverage the global properties of local features \cite{DCM,SOGM,Lorenzo2008ECCV,BBHomo}. In most practical applications, a simple yet effective method is perform epipolar check after NN matching that estimate essential matrix between two images based on NN matched local features, and reject mismatched features being contrary to the underlying epipolar constraints formatted as essential matrix \cite{orbslam,orbslam2,orbslam3}. But if mismatched local features were in the majority, the estimated essential matrix will be wrong and this solution will fail. Thus, we usually integrated this solution into a RANSAC loop \cite{ransac} to obtain estimated model with most inliers and use the model to reject mismatches. Recently, AdaLAM \cite{adalam} has been proposed to detects inliers by searching for significant local affine patterns in image correspondences. These traditional hand-crafted matching methods have large running latency and require costly computational consumption due to complex rules and iterative formulation, and the performance will degrade if the underlying model between local features in real scenarios differs from the pre-defined model. In \cite{VPS}, Sattler \textit{et al.} proposed a novel vocabulary-based prioritized matching step that enables to first consider features more likely to yield 2D-3D matches and to stop searching matches as soon as enough matches have been found, making local feature matching much more efficient and effective.
}

As for DL-based local feature matching methods, some algorithms explore semantic information to match the semantic key points in different instances of the same category of objects \cite{DeepSemanticFM,Yu2018HierarchicalSemanticIM}. 
Some works regard local feature matching as a point set learning problem and develop putative matches filtering networks \cite{good_correspondence,NMNet,OANet,ACNe}. They commonly organize the putative matches of key points as 4D quads and feed the 4D quads into network to estimate scores of each putative match. False matches are discarded by a threshold on the scores. These methods only consider the position information of the key points and neglect the descriptor information of local features, thus remaining room for improvements.

In 2019, SuperGlue \cite{superglue} was proposed as an E2E learning-based approach for local feature matching and achieved impressive success. SuperGlue \cite{superglue} considers both key points and descriptors of local features and learns their matches with a Graph Neural Network (GNN). Each local features are modelled as a node in the graph and the descriptors are enhanced by all the features in both images by an attentional GNN layer so that matching will be reliable. Later, Zhao \textit{et al.} \cite{IMUPrior} exploited motion estimation from IMU integration and use the estimation as a spatial distribution prior of matched local features between images. With the assistance of such a prior, the network can achieve comparable accuracy with less GNN layers, boosting the real-time performance. More recently, LightGlue \cite{lindenberger2023lightglue} improves the inference speed by enabling the model prune unmatchable local features and predict whether further computation is required.

\begin{figure}[!h]
    \centering
    \includegraphics[width=0.97\linewidth]{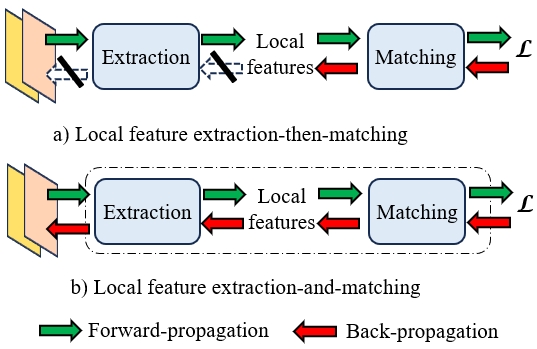}
    \caption{The diagrams of a) local feature extraction-then-matching and b) extraction-and-matching methods.}
    \label{fig:feature-match-diagram}
\end{figure}

{
\vspace{4pt}
\setlength{\parindent}{0cm}
\textbf{Summary:}
\label{sec:extract-then-match summary}
The local feature extraction-then-matching scheme has worked well in most cases and supported high-precision VL-MRL for a long time. However, there remains a fundamental shortcoming that they cannot match well if the extracted local features were extremely bad. The supervision on matches can only optimize matching module but cannot make any effort on the extraction module, as shown in Fig. \ref{fig:feature-match-diagram} a). To solve this problem, joint local feature extraction-and-matching solution, or called detector-free matching, is proposed.
}

\subsection{Joint Local Feature Extraction and Matching}
\label{sec:local feature extraction and matching}

Joint local feature extraction-and-matching methods integrate feature extraction and matching module in a unified DNN model, and they directly estimate dense local feature matches so that the two modules can be jointly optimized in an E2E way, as shown in Fig. \ref{fig:feature-match-diagram} b). Early proposals in this area adopt CNN model, \textit{e.g.}, NCNet \cite{NCNet}, Sparse NCNet \cite{SparseNCNet}, and DRC-Net \cite{DRCNet}. Although all the possible matches should be considered in the CNN of these works, the receptive field of convolution is still limited to neighborhood area, making matching sided and unreliable.

To solve the problem, in recent three years, some works begin to adopt global attention mechanism by Transformer \cite{transformer} to achieve global consensus between matches. 
As the first proposal in this line, COTR \cite{COTR} use a Transformer-based model that queries a local feature in one image and estimates the matched local feature on another image. For efficiency, COTR \cite{COTR} apply the network by recursively zooming in around the estimations. LoFTR \cite{LoFTR} proposes to perform dense matching in a coarse-to-fine framework, that it firstly obtains confidence matrices for patch-level coarse matching and then refines matched points positions at pixel level. Both steps using self- and cross-attention layers by Transformer \cite{transformer} to enhance the representative ability of feature descriptors. It achieves accurate matching performance even in textureless region, effectively reducing the requirements of scene texture in VL-MRL methods. 
As follow-up methods, some methods make efforts to improve the matching accuracy and efficiency \cite{Tang2022QuadTreeAF,Chen2022ASpanFormerDI,Liao2023TKwinFormerTK}.


Existing matching methods choose to ignore the occluded area, and they will fail when there is no overlap region between $\mathcal{I}_q$ and $\mathcal{I}_r$. To overcome this limitation, $\mbox{OCC}^2\mbox{Net}$ \cite{Fan2023Occ2NetRI} models occlusion relations using 3D occupancy, enabling local feature matching for occluded regions.

{
\vspace{4pt}
\setlength{\parindent}{0cm}
\textbf{Summary:}
\label{sec:extract-and-match summary}
The joint extraction-and-matching scheme breaks through the limits of extraction-then-matching schemes that good matching relies on excellent local features, and achieves impressive performance, making VL-MRL methods more accurate. But it requires the scene map additionally saves $\mathcal{I}_r$, enlarging the storage requirements of scene map.
}

\subsection{Pose Solver}
\label{sec:pose solver}

Given the camera instrinsic $K$ and camera pose $\textbf{x}^\mathcal{G}_q$, the 2D camera projection $\textbf{p}^{q}_{i}=\left[u,v\right]$ from 3D visual landmarks $\textbf{m}_i=<\textbf{P}_i,*>, \textbf{P}_i=\left[X^\mathcal{G}_i,Y^\mathcal{G}_i,Z^\mathcal{G}_i\right]$ to image plane $\mathcal{I}_q$ can be formatted as:
\begin{equation}
    \begin{split}
        \label{equ:pnp1}
        \left[\begin{array}{c}
                \textbf{p}^{q}_i \\
                1
            \end{array} \right] & = \frac{1}{Z^\mathcal{C}_i} K \left[X^\mathcal{C}_i~Y^\mathcal{C}_i~Z^\mathcal{C}_i\right]^T \\
        &= \frac{1}{Z^\mathcal{C}_i} K \left[\textbf{R}^\mathcal{C}_\mathcal{G}~|~\textbf{t}^\mathcal{C}_\mathcal{G}\right]\left[X^\mathcal{G}_i~Y^\mathcal{G}_i~Z^\mathcal{G}_i\right]^T \\
        &= \frac{1}{Z^\mathcal{C}_i} K \left(\textbf{x}^\mathcal{G}_\mathcal{C}\right)^{-1} \textbf{P}_i
    \end{split}
\end{equation}
where ${\{\cdot\}}^\mathcal{C}$ denotes points in camera coordinate $\mathcal{C}$. And PnP problem is to estimate $\hat{\textbf{x}}^\mathcal{G}_q$ given $K$ and $n$ 2D-3D correspondences $\{\textbf{p}^{q}_i,\textbf{P}_i\}^{n}_{i=1}$. This is a typical problem in 3D vision. Direct Linear Transformation (DLT) \cite{DLT} was first developed by photogrammetrists to solve PnP problem and then introduced to computer vision community. In the last two decades, PnP solvers have been well studied. 
The minimum number of correspondences to solve PnP problem is reduced from 5 \cite{P5P}, 4 \cite{P4P}, to 3 \cite{p3p2,p3p1}. 

Generally, the PnP solvers can be categorized as non-iterative and iterative. Early non-iterative PnP solvers are generally computationally expensive, \textit{e.g.}, $O(n^5)$ \cite{Quan1999TPAMI}. The EPnP solution \cite{Lepetit2009EPnPAA} effectively reduces the computation complexity to $O(n)$ by representing 3D point coordinates as a linear combination of four control points, which is widely utilized in many MRL solutions. In addition to EPnP \cite{Lepetit2009EPnPAA}, some other non-iterative PnP solvers with $O(n)$ complexity are later proposed \cite{RPnP,OPnP,SRPnP,EOPnP} but they are polynomial solvers compared to linearization-based EPnP \cite{Lepetit2009EPnPAA}. Instead, iterative methods solve PnP problems in an iterative manner, beginning from an appropriate initial value \cite{LHM,FP,SQPnP}. The non-iterative solutions are employed to provide an initial value, and iterative algorithms are used to refine the estimation \cite{Lepetit2009EPnPAA}.

The PnP problem can also be seen as a non-linear least square problems with regard to re-projection error so that camera pose can be solved by Bundle Adjustment (BA) \cite{orbslam3}. From Eqn. \ref{equ:pnp1}, the re-projection error is defined as:
\begin{equation}
    \begin{split}
        \label{equ:pnp2}
        e_i =\textbf{p}^{q}_i - (\frac{1}{Z^\mathcal{C}_i} K \left(\textbf{x}^\mathcal{G}_\mathcal{C}\right)^{-1} \textbf{P}_i)[:2]
    \end{split}
\end{equation}
To estimate optimal camera pose, we need to minimize the total re-projection errors:
\begin{equation}
    \begin{split}
        \label{equ:pnp3}
        \hat{\textbf{x}}^\mathcal{G}_q = \hat{\textbf{x}}^\mathcal{G}_{\mathcal{C}} = \mathop{\arg\min}\limits_{\textbf{x}^\mathcal{G}_{\mathcal{C}}} \frac{1}{2} \sum^n_{i=1} e^2_i
    \end{split}
\end{equation}
And the function can be iteratively done by non-linear optimization methods like Gauss-Newton (GN) method and Levenberg-Marquardt (LM) method. In ORB-SLAM3 \cite{orbslam3}, the VL-MRL after tracking lost or kidnapping is implemented by first using MLPnP solver \cite{Urban2016MLPnPA} to estimate an initial result and then refining with BA optimization.

{
\vspace{4pt}
\setlength{\parindent}{0cm}
\textbf{Summary:}
Given accurate 2D-3D correspondences, PnP is a well-established problem and can be excellently solved. Pose can be precisely solved. However, in most real-world cases, 2D-3D correspondences usually have many or even predominant mismatches, challenging the pose estimation procedure. The PnP solver is always integrated into a RANSAC loops \cite{ransac} to estimate more robustly and precisely \cite{hloc2,orbslam3}.
}

\subsection{Further Improvements}
\label{sec:further improvements}

Except mentioned above methods, there are many attempts to improve VL-MRL methods in other perspectives. Here we introduce some works with noticeable improvements.

{
\vspace{4pt}
\setlength{\parindent}{0cm}
\textbf{Cross Descriptor matching \cite{cross-desc}:} Traditional VL-MRL methods supposes to use the same local features in mapping and localization stage, which blocks the development of new local feature algorithm. For example, if we used SIFT \cite{sift} in mapping while HardNet \cite{hardnet} in localization, traditional solution fails. In \cite{cross-desc}, the authors overcame this problem by proposing to translate different type of descriptors into other descriptors or joint embeddings so that different descriptors can be indirectly matched.
}

{
\vspace{4pt}
\setlength{\parindent}{0cm}
\textbf{Dense CNN Matching \cite{InLoc,PoseVerification}:} Indoor environments have large textureless areas where sparse local feature methods detect very few features and the extracted features are clustered in small, well-textured regions in the image. In this case, pose cannot be stably solved. InLoc \cite{InLoc} and \cite{PoseVerification} use MS dense CNN features for feature matching. 
}

\begin{figure*}[!t]
    \centering
    \includegraphics[width=0.97\linewidth]{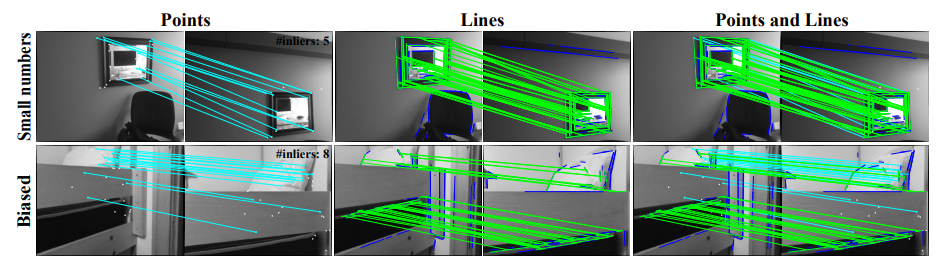}
    \caption{Points and lines are complementary for better localization \cite{PL-Loc}.}
    \label{fig:PL}
\end{figure*}

{
\vspace{4pt}
\setlength{\parindent}{0cm}
\textbf{Line Feature \cite{PL-Loc}:} Along with point-wise local features for matching, line-wise local features may futher help VL-MRL methods, especially in human-made environment. In \cite{PL-Loc}, authors first presented a novel line segment descriptor and suggested the line feature can be complementary constraints to traditional point feature, as show in Fig. \ref{fig:PL}, thereby boosting localization performance on extreme scenarios where point features are biased or sparse.
}

{
\vspace{4pt}
\setlength{\parindent}{0cm}
\textbf{VL-MRL without SfM \cite{meshloc}:} Traditional VL-MRL need to reconstruct visual landmark map using SfM \cite{colmap} in offline mapping stage and the visual landmark map is generally 3D points, which is computational and memorial costly. MeshLoc \cite{meshloc} explores a more flexible alternative based on dense 3D meshes that does not require local features matching between $\mathcal{I}_r$ to build the scene map in mapping stage. During localization stage, it still perform local feature matching on 2D images and estimate pose by solving PnP problem, but the only difference is $\mathcal{I}_r$ is online rendered by 3D mesh model.
}

{
\vspace{4pt}
\setlength{\parindent}{0cm}
\textbf{Map Squeeze \cite{SceneSqueezer}:} Storing pre-built visual landmark map can be prohibitively expensive for large-scale environments, especially on mobile devices with limited storage and communication bandwidth. In \cite{SceneSqueezer}, authors designed a novel framework, SceneSqueezer, to compresses a scene while still maintaining localization accuracy. The scene $\mathcal{M}$ is compressed in three aspects: 1) $\mathcal{I}_r$ for mapping are clustered; 2) the visual landmarks in each cluster are pruned; 3) the descriptors of the selected visual landmarks are further compressed. By applying SceneSqueezer \cite{SceneSqueezer}, the size of map is reduced from 7828 MB to 31 MB in Aachen Day-Night dataset \cite{benchmark6dof} while achieving comparable localization performance.
}

{
\vspace{4pt}
\setlength{\parindent}{0cm}
\textbf{Pose Verification \cite{InLoc,PoseVerification} and Correction \cite{PoseCorrection}:} Some VL-MRL methods estimate a set of candidate poses for $\mathcal{I}_q$, a pose selection procedure needs to be conducted. 
Given an estimated \textit{query} pose, InLoc \cite{InLoc} utilizes the 3D structure of indoor scene to explicitly renders a virtual image and measure its similarity to $\mathcal{I}_q$ as a score of \textit{query} poses, which is called pose verification.
In \cite{PoseVerification}, Taira \textit{et al.} improved the procedure by integrating semantic constraint and trainable pose verification network. 
In \cite{PoseCorrection}, Hyeon \textit{et al.} proposed pose correction to reorganize local features observed from the estimated pose so that more reliable candidate poses can be provided to pose selection.
}

{
\vspace{4pt}
\setlength{\parindent}{0cm}
\textbf{Open-sourced Toolbox \cite{hloc2,Kapture,xrlocalization}:} To facilitate the development of VL-MRL researches, many great works have open-sourced the overall localization pipeline for the community, \textit{e.g.}, HLoc\footnote{https://github.com/cvg/Hierarchical-Localization}, Kapture-Localization\footnote{https://github.com/naver/kapture-localization}, and XRLocalization\footnote{https://github.com/openxrlab/xrlocalization}. By leverage these toolboxes, researchers can quickly valid their new contributions or evaluate their new proposals without any extra efforts to build localization pipeline.
}

\section{Point Cloud Map}
\label{sec:geometry}

Visual landmark map has a large requirement of storage for saving high-dimensional descriptors. The sensitivity of appearance also challenges mapping with monocular camera and affect the reconstruction accuracy of visual landmark map. Instead, LiDAR can provide direct and precise 3D perception of scene map geometry, and the point cloud map built by high-precision LiDAR always be illumination-invariant and commonly more accurate than visual map. And point cloud map only need to save 3D position of point cloud, alleviating the storage burden. The point cloud map can be represented as $\mathcal{M}=\{\textbf{m}_i\}$ where $\textbf{m}_i$ is a 3D point in global frame $\mathcal{G}$, $\textbf{m}_i=<\textbf{p}^\mathcal{G}_i, \textbf{l}_i>$, $\textbf{P}^\mathcal{G}_i$ is its 3D coordinates and $\textbf{l}_i$ is its intensity (intensity sometimes is missing). In practical usage, the 3D points are usually used in ``clip'' format, that is, given a virtual viewpoint $\textbf{x}^\mathcal{G}_v$, the 3D point observed in this view can be clustered in a set $\mathcal{P}^\mathcal{G}_v=\{\textbf{P}_i,\textbf{l}_i\}$, so $\mathcal{M}$ is changed to $\mathcal{M}=\{\textbf{m}_i\}$ where $\textbf{m}_i$ is a clip of 3D point and its associated viewpoint $\textbf{m}_i=<\mathcal{P}^\mathcal{G}_v,\textbf{x}^\mathcal{G}_v>$.
Since the wide deployment of LiDAR localization is hampered by the huge cost of high-precision LiDAR, localizing camera in LiDAR point cloud map (named as Image-to-Point cloud (I2P) localization for simplicity) is a newly emerged trend in MRL. However, the appearance information of camera and the geometry information of LiDAR is in different modality, the inherit difficulties in cross-modal data association challenges I2P Localization. In this section, we review cross-modal I2P localization using traditional geometry rules and using learning-based methods.

\begin{figure*}[!t]
    \centering
    \includegraphics[width=0.97\linewidth]{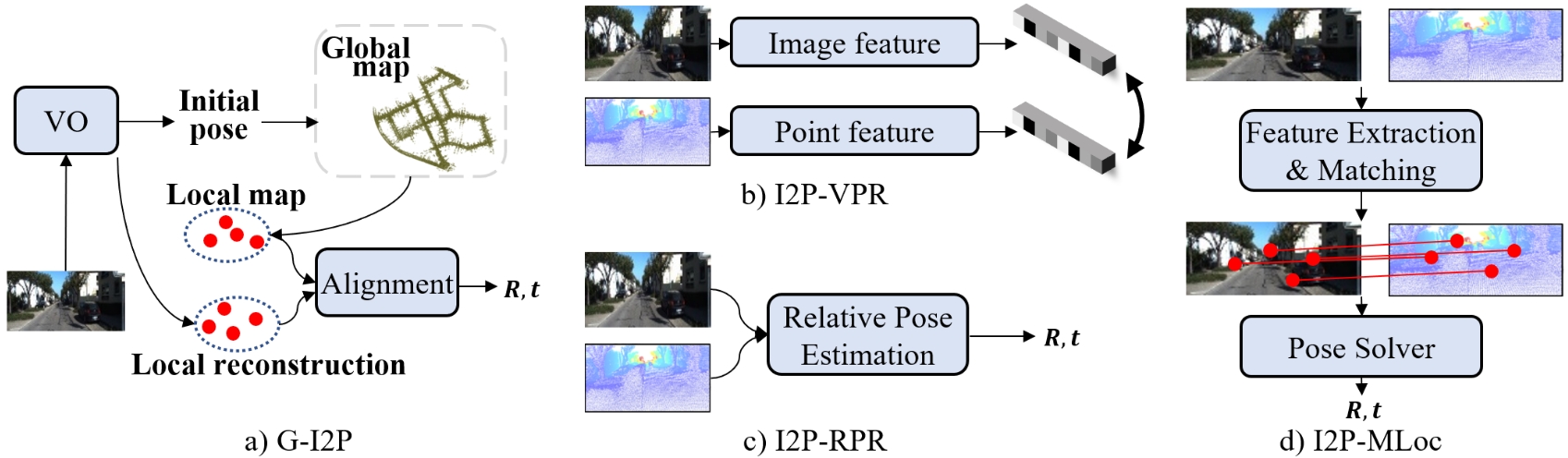}
    \caption{The diagram of four typical cross-modal I2P localization methods. including geometry-based a) G-I2P \cite{mono_i2p_vo}, and learning-based b) I2P-VPR, c) I2P-RPR, and d) I2P-MLoc.}
    \label{fig:i2p}
\end{figure*}

\subsection{Geometry-based Cross-modal Localization}
\label{sec:geo-based i2p}
Geometry-based I2P (G-I2P) methods localize camera in LiDAR map by traditional geometry rules. In the G-I2P researches, stereo camera is a natural pick because we can easily lift 2D image pixels to 3D point clouds by stereo matching, which makes the alignment between camera and point cloud map much easier \cite{stereo_i2p}. Instead, localizing monocular camera images in 3D point cloud maps is inherently more challenging, as they lack any absolute depth or 3D information. For this, some methods align $\mathcal{I}_q$ to $\mathcal{M}$ by lifting 2D visual measurements to 3D local construction \cite{mono_i2p_vo}, digging geometry features \cite{mono_i2p_line,Wolcott2014IROS}, or modelling 3D structure in other advanced way \cite{mono_i2p_SDF,dsl,Huang2020GMMLocSC}. 

As an early and successful attempt, Caselitz \textit{et al.} \cite{mono_i2p_vo} proposed to reconstruct local visual landmarks by a monocular Visual Odometry (VO) \cite{orbslam} and simultaneously estimate initial pose. Using initial pose $\hat{\textbf{x}}^{\mathcal{G}}_{init}$, a local map is retrieved and cropped from global point cloud map. The resulting local reconstruction is aligned to local map in an ICP scheme so that a 7 DoF transform (including 6 DoF relative pose $\hat{\textbf{x}}^{init}_{v}$ with regard to local map and 1 DoF scale) can be estimated as shown in Fig. \ref{fig:i2p} a). MRL is achieved by $\hat{\textbf{x}}^{\mathcal{G}}_{v}=\hat{\textbf{x}}^{\mathcal{G}}_{init}*\hat{\textbf{x}}^{init}_{v}$.
Then in \cite{Wolcott2014IROS}, authors estimate pose by maximizing the normalized mutual information between real \textit{query} image $\mathcal{I}_q$ and synthetic images rendered by LiDAR intensity. Although identifying matched features across different modalities is proven challenging, the incorporation of matched 2D-3D line-level features between query image and point cloud map has been demonstrated to aid in this objective in \cite{mono_i2p_line}. In DSL \cite{dsl}, Ye \textit{et al.} introduced the surfel constraints of point clouds into the direct photometric error and estimated poses by a tightly-coupled BA framework. Huang \textit{et al.} first modelled the dense structure as an Euclidean Signed Distance Field (ESDF) in \cite{mono_i2p_SDF} and then improved by modelling the prior distribution of the point cloud map by Gaussian Mixture Model (GMM) \cite{GMM} in \cite{Huang2020GMMLocSC}. 

{
\vspace{4pt}
\setlength{\parindent}{0cm}
\textbf{Summary:} Many G-I2P methods achieve considerable successes. With the absolute pose constraint by G-I2P and relative temporal constraint by VO, G-I2P methods can alleviate the accumulated drift problem of long-term VO and smoothen noisy estimation of absolute localization, achieving consistent and accurate pose estimation.
But these works are limited to be coupled in a VO pipeline to obtain initial pose and hard to re-localize with only one monocular image. The monocular G-I2P solution is still an unsolvable problem.
}

\subsection{Learning-based Cross-modal Localization}
\label{sec:learning-based i2p}

{
\vspace{4pt}
\setlength{\parindent}{0cm}
\textbf{Cross-modal Visual Place Recognition (I2P-VPR):} Recently, scholars have begun to explore DL-based methods to achieve I2P localization with only one monocular image. First, many researches adopt GF-VPR method into this cross-modal I2P localization task and propose I2P-VPR algorithms. I2P-VPR methods are very similar to GF-VPR methods, but its \textit{reference} ``frames'' $\mathcal{I}_{r(t)}$ is changed to local clips of point cloud map $\mathcal{P}^\mathcal{G}_{r(t)}$. Given the \textit{query} image $\mathcal{I}_q$, a I2P-VPR algorithm can be written as a two-step pipeline, that is,
}
\begin{itemize}
    \item[1] represent $\mathcal{I}_q$ and all the \textit{reference} point cloud clips $\mathcal{P}^\mathcal{G}_{r(t)}$ in $\mathcal{M}$ as features:
        \begin{equation}
            \begin{split}
                \label{equ:i2p1}
                \mathcal{F}^{q}&=\mbox{ImFeat}\left(\mathcal{I}_q\right)~~\mbox{for \textit{query}} \\
                \mathcal{F}^{r(t)}&=\mbox{PcFeat}\left(\mathcal{P}^\mathcal{G}_{r(t)}\right)~~\mbox{for \textit{reference}}
            \end{split}
        \end{equation}
    \item[2] retrieve the best-matched point cloud clip based on the similarity of features:
        \begin{equation}
            \label{equ:i2p2}
            \hat{\textbf{m}}=\mathop{\arg\max}\limits_{\mathbf{m}_i} P(\{\mathcal{F}^{q}, \mathcal{F}^{i}~\mbox{of}~\textbf{m}_i | \textbf{m}_i \in \mathcal{M})
        \end{equation}
\end{itemize}
Then, the pose of $\mathcal{I}_q$ can be approximated as the associated pose of $\hat{\textbf{m}}$.

For this problem, one straightforward solution is jointly training a 2D CNN $\mbox{ImFeat}(\cdot)$ for images and a 3D DNN $\mbox{PcFeat}(\cdot)$ for point cloud clips to create matchable feature between cross-modal data \cite{2d3d_embed}. However, this approach does not generalize well to unseen environments. \cite{Lukas2021ICRA_Spherical} proposes to project images and point clouds into unit spheres and extract features through sphere CNN, which requires multiple images as input. Towards I2P localization robust to inconsistent environmental conditions, i3dLoc \cite{i3dloc} matches equirectangular images to the 3D range projections of point cloud map. In i3dLoc \cite{i3dloc}, authors concludes visual condition-related information of scenes makes robust matching challenging, so we should eliminate condition-related features and extract condition-invariant 3D geometry features from visual inputs for matching to point cloud map only including geometry information. Therefore, a Generative Adversarial Network (GAN) is designed to extract features matchable to point cloud map. To improve performance of I2P-VPR, i3dLoc \cite{i3dloc} also designs a spherical convolution network to learn viewpoint-invariant features. Similarly, \cite{AE-CrossModal} focuses on correlating the information of 360-degree spherical images to point clouds. Attention mechanism is applied to let the network capture the salient feature for comparing images and point clouds. I2P-Rec \cite{Li2023I2PRecRI} provides a new baseline for I2P-VPR that it leverages on depth estimation networks to recover point clouds from images so that cross-modal data is converted into the same modality, \textit{i.e.}, Bird-Eye-View (BEV) images. Using the BEV image as an intermediate representation, very simple global feature network (CNN encoder followed by NetVLAD layer \cite{netvlad}) trained by few training data can achieve SOTA localization performance.

{
\vspace{4pt}
\setlength{\parindent}{0cm}
\textbf{Cross-modal Relative Pose Regression (I2P-RPR):} After obtaining matched point cloud clip $\mathcal{P}^\mathcal{G}_r~\mbox{of}~\textbf{m}_i$ and its pose $\textbf{x}^\mathcal{G}_r$, many works further refine \textit{query} pose by I2P-RPR that is similar to image-to-image RPR in Sec. \ref{sec:RPR}, that is, estimate relative pose between $\mathcal{I}_q$ and $\mathcal{P}^\mathcal{G}_r$:
}
\begin{equation}
    \label{equ:i2p3}
    \hat{\textbf{x}}^r_q = \mbox{I2P-RPR}(\mathcal{I}_q,\mathcal{P}^\mathcal{G}_r)
\end{equation}
so a fine absolute pose of $\mathcal{I}_q$ can be calculated by:
\begin{equation}
    \label{equ:i2p4}
    \hat{\textbf{x}}^\mathcal{G}_q \leftarrow \textbf{x}^\mathcal{G}_r * \hat{\textbf{x}}^r_q
\end{equation}

In this field, Cattaneo \textit{et al.} presented CMRNet \cite{cmrnet} that calculates the cost volume between the image feature and LiDAR feature and then regresses the pose of monocular camera relative to the LiDAR map. The LiDAR map provides projected depth image for the network as input.
Built upon this, Chang \textit{et al.} compressed the LiDAR map to reduce map size by 87-94$\%$ while achieving comparable accuracy \cite{hypermap}. Such I2P-RPR networks simply utilize stacked convolution layer or fully connection layer as pose regressor and estimate pose in one shot that cannot fully exploit pose-related information from image-point cloud cost volumes. Therefore, POET \cite{miao2023poses} converts pose to high-dimensional features as queries in Transformer \cite{transformer} and iteratively optimizes pose within a Transformer-based pose regressor, achieving improved localization accuracy.

However, I2P-VPR and I2P-RPR methods are hard to achieve centimeter-level accuracy on large scale scene. So many methods begin to explore 2D-3D matches between images and point cloud map, which is technically similar to VL-MRL methods.

{
\vspace{4pt}
\setlength{\parindent}{0cm}
\textbf{Cross-modal Matching-based Localization (I2P-MLoc):} 2D3D-MatchNet \cite{2d3dmatchnet} is one of the earliest works focusing on image-to-point cloud matching for robot localization. The 2D and 3D key points are obtained by SIFT \cite{sift} and ISS \cite{ISS} respectively. Then, a neural network is introduced to learn the descriptors for cross-modal key points. Finally, EPnP \cite{Lepetit2009EPnPAA} is adopted to estimate the pose of the image relative to the point cloud based on 2D-3D correspondences.
Built upon CMRNet \cite{cmrnet}, CMRNet++ \cite{cmrnet++} turn to regress displacements between images and projected depth image of point cloud map instead of directly regressing pose. CMRNet++ \cite{cmrnet++} achieves improved performance with significant margin, indicating that matching-based methods should be a more reasonable choice than regression-based alternates for I2P localization.
DeepI2P \cite{deepi2p} splits the image-to-point cloud matching problem into a classification problem and an optimization problem. A cross-modality neural network is adopted to classify whether the points fall into the image frustum and the classification results are formatted as a cost function to measure how close the estimated pose to real one in optimization.
CorrI2P \cite{CorrI2P} designs a cross-modality network to extract the image-to-point cloud overlapping region and corresponding dense descriptors for the image and point cloud. CorrI2P \cite{CorrI2P} constructs dense image-to-point cloud correspondences and uses iterative RANSAC-based EPnP \cite{Lepetit2009EPnPAA,ransac} to estimate the relative pose.
EFGHNet \cite{EFGHNet} adopts the divide-and-conquer strategy to divide the image-to-point cloud matching problem into four separate sub-networks. These sub-networks are responsible for the horizon and ground normal alignments, rotation estimation, and translation estimation, respectively.
These mentioned above methods divide the I2P-MLoc into two steps for robot localization, \textit{i.e.}, matching and localization. The separation makes the two steps separately optimized and thus not able to refine the error of the previous step. Thus, I2D-Loc \cite{I2DLoc} apply BPnP module \cite{BPnP} to calculate the gradients of the back-end PnP-based pose estimation process, enabling the model to be trained end-to-end with the supervision on estimated pose.
}

More recently, I2PNet \cite{I2PNet} utilizes a coarse-to-fine architecture to accurately regress pose. By combining both advantages of I2P-RPR and I2P-MLoc methods, I2PNet \cite{I2PNet} could achieve centimeter-level localization accuracy.

Some other works add visual features in point cloud map, which is quite similar to visual landmark map in Sec. \ref{sec:visual} but the visual features of point cloud in map are online extracted during localization without the need of offline storage. \cite{NRE} introduces the Neural Re-projection Error (NRE) as a substitute for re-projection error in PnP problem. Given $\mathcal{I}_q$ and $\mathcal{P}^\mathcal{G}_r$ in map with $\mathcal{I}_r$, the method extracts dense CNN features from both images. The sparse descriptors of 3D point clouds are sampled from $\mathcal{I}_r$ based on image projection as Eqn. \ref{equ:pnp1}. For each 3D point, \cite{NRE} computes dense loss maps and minimizes the loss with respect to the network parameters. The proposed NRE in \cite{NRE} is differentiable not only \textit{w.r.t.} the camera poses but also \textit{w.r.t.} the descriptors, so camera pose and descriptors can both be E2E optimized. PixLoc \cite{pixloc} is another work in this field. PixLoc \cite{pixloc} extracts multi-level features with pixel-wise confidences for $\mathcal{I}_q$ and $\mathcal{I}_r$. The LM optimization aligns corresponding features according to the 3D points in the map, guided by the confidences, from the coarse to the fine level.

{
\vspace{4pt}
\setlength{\parindent}{0cm}
\textbf{Summary:} The inherit difference between cross-modal data make I2P localization hard. Benefit from high-dimensional CNN features, matching across cross-modal data becomes possible. In general, I2P-VPR only serves to retrieve \textit{reference} point cloud clip and provides an coarse pose estimation. I2P-RPR and I2P-MLoc could provide more precise pose estimation. I2P-RPR avoid explicit feature matching and pose solver, the training cost is cheap but its localization accuracy is limited. I2P-MLoc performs best in cross-modal localization task, showing good prospects for developments.
}

\begin{figure}[!h]
    \centering
    \includegraphics[width=0.97\linewidth]{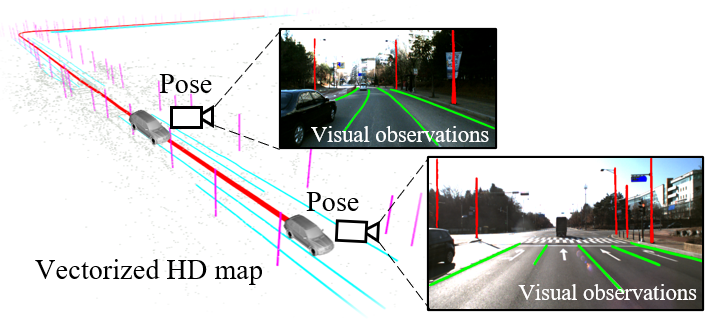}
    \caption{A diagram of MRL method with vectorized HD map.}
    \label{fig:hdloc}
\end{figure}

\section{Vectorized High-Definition Map}
\label{sec:hdmap}

In AD task, the compactness and accuracy are two vital required characteristics of scene map. Therefore, the most widely used map in AD area has been HD Map since the its first appearance in late 2010. Commonly, the HD Map is created by mobile mapping system equipped with high-precision sensors including LiDAR, RTK and IMU at centimeter-level precision \cite{HDMap_review}. The localization feature in the HD map includes dense point cloud and sparse map element. The dense point cloud in HD Map is similar to the raw point cloud map in Sec. \ref{sec:geometry}. In this section, we focus on the sparse map elements usually represented as vectors with semantic labels. These elements correspond to road element or signatures in the real world, \textit{e.g.}, lighting poles, road markers, and lane lines. Such a semantic element-based map representation is much lighter than other scene map like point cloud map and visual landmark map while maintains highly detailed road elements useful to AD task \cite{HDMap_review}.
MRL with HD map (HD-MRL) is concluded as an effective solution for mass-produced vehicles. The fundamental formulation of HD-MRL methods involve detecting semantic map elements from images and then estimating pose by aligning the detected 2D element with their corresponding 3D element in the HD map.

Generally, a HD map can be denotes as $\mathcal{M}=\{\textbf{m}_i\}$ where each map element $\textbf{m}_i$ with its semantic category $s_i$ is modeled as a set of 3D control points $\textbf{m}_i=<\{\textbf{P}_{i(j)} \in \mathbb{R}^3 \}_{j=1:N_i}, \textit{s}_i>$ sampled uniformly in the 3D space for a unified representation, here $N_i$ as the amount of control points of $\textbf{m}_i$. Given the camera pose $\textbf{x}^\mathcal{G}_q$, the 3D points $\{\textbf{P}_{i(j)}\}$ of $\textbf{m}_i \in \mathcal{M}$ can be projected into the image space as Eqn. \ref{equ:pnp1}, obtaining 2D projection $\{\textbf{p}^{q}_{i(j)}\}$. The HD-MRL methods is to find an optimal camera pose $\hat{\textbf{x}}^\mathcal{G}_{q}$ which can minimize a defined cost model $d(\cdot, \cdot)$ between the projected HD map element points and their corresponding observations:
\begin{equation}
    \label{equ:hdmap1}
    \hat{\textbf{x}}^\mathcal{G}_{q} = \mathop{\arg\min}\limits_{\textbf{x}^\mathcal{G}_q} \sum_{\textbf{m}_i \in \mathcal{M}} \sum^{N_i}_{j=1} d\left( \textbf{z}^{q}_{i(j)},\textbf{p}^{q}_{i(j)} \right)
\end{equation}
where $\textbf{z}^{q}_{i(j)}$ is the observation of $j$-th point of $i$-th element $\textbf{m}_i$ in image plane $\mathcal{I}_q$.

From the basic formulation defined above, HD-MRL methods are usually integrated into an online localization system with multiple sensor fusion for acquiring an initial camera pose, which is similar to G-I2P, since HD map elements can be seen as sets of 3D point cloud with semantic information. 
In this field, Pink \textit{et al.} \cite{Oliver2008CVPRW} built a lane-level map by aerial images, and then matched the lane markings in the \textit{query} image with the pre-built map using ICP algorithm. 
However, the sparsity of HD map elements in real-world scenarios restricts the available information for pose estimation, blocking the wide usage of HD-MRL methods. 
To solve this problem, multiple sensor fusion often be used in Extended Kalman Filter (EKF) frameworks where IMU and GPS are utilized for continuous localization \cite{Tao2013IROS,Cai2018Sensors}. 
For the same purpose, recent advances in HD-MRL begin to integrate relative pose constraint by visual feature tracking in VO \cite{vins-mono} and visual(-inertial) SLAM \cite{orbslam,orbslam2,orbslam3} so that tracked visual features can be considered as complementary localization information to the HD map elements. 
In this area, some solutions combine these two pieces of information in a loosely coupled manner. They first perform HD-MRL and VO separately and then fused by Kalman Filter \cite{Julius2014IV} or sliding window-based state estimator \cite{MLVHM}. However, such a loosely-coupled strategies require the HD map elements to be minimally self-sufficient for pose estimation, which does not fundamentally address the failure case caused by sparsity of HD map element. So, Wen \textit{et al.} \cite{Wen2020IV,TM3Loc} followed this way to combine VO and HD-MRL but in a tightly-coupled framework with sliding window-based optimization. The optimization target of pose estimation considers both relative constraints of VO and absolute constraint of HD-MRL, enabling consistent and accurate localization even when HD map elements are insufficient in current observation.

\begin{figure}[!t]
    \centering
    \includegraphics[width=0.97\linewidth]{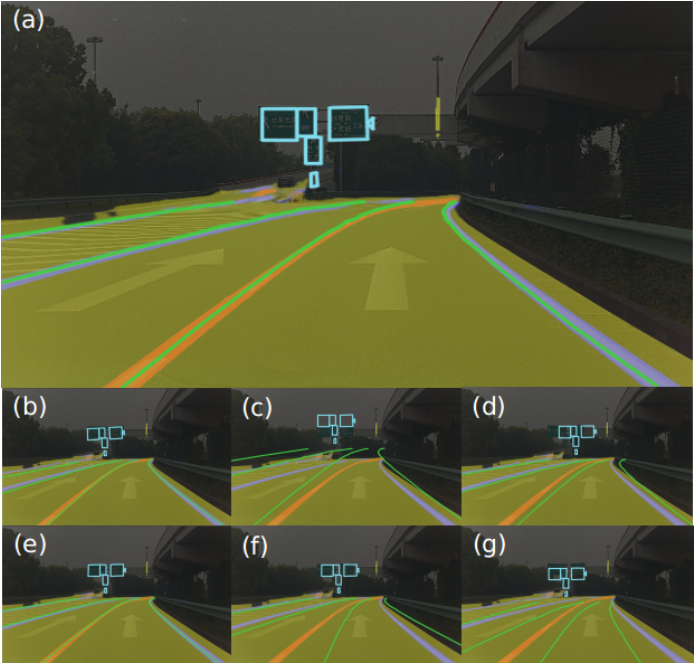}
    \caption{Examples of alignment between HD map and image semantic segmentation \cite{Guo2021IROS}. (a) An accurate alignment; (b)-(g) Alignment results under perturbation of freedom: roll, pitch, yaw, x, y and z. Semantic segmentation: freespace (yellow), lane markings (organ and purple). Projection from HD map: lane markings (green) and signboard (cyan).}
    \label{fig:hdmap}
\end{figure}

Also, the depth ambiguity of monocular camera makes 2D-3D association between observed map elements in \textit{query} image and the ones in HD map a technically challenging problem. 
For precisely matching map elements, we need to project the map elements in \textit{query} image and HD map into a unified space. 
As the 3 DoF motion of vehicles using HD-MRL is generally stuck to the ground, one common strategy is to transform the detected map elements into the BEV using Inverse Projection Mapping (IPM) algorithms \cite{Derenzi2018TITS,Deng2019SensorsJ}. 
This strategy follows an ideal assumption that the road is flat, so it will suffer a serious disruption if there was an indispensable vibration on camera tilt angle with respect to the ground plane \cite{Jang2022JAS}. 
Some other works turn to project map elements from HD map into \textit{query} image plane given initial \textit{query} poses and perform matching on 2D image space \cite{Xiao2018TISC,Wen2020IV}, as shown in Fig. \ref{fig:hdmap}. 
Due to the perspective effect of the monocular camera model, the shape of map elements will differ from their original 3D shape in HD map. 
So, the parameterization of map elements and the definition of cost model $d(\cdot, \cdot)$ need to be investigated. In \cite{Xiao2018TISC,Wen2020IV}, authors simply modelled land boundaries as straight lines so that a point-to-line cost model between observed and projected map elements can be used to solve poses.
Liao \textit{et al.} \cite{Liao2019CoarseToFineVL} instead utilized the pole-like landmark features and represented them as straight lines in the image plane, and localized by a particle filter. 
For fully using map elements in the scene, Guo \textit{et al.} \cite{Guo2021IROS} used different post-processing methods for semantic segmentation of different elements in HD map. 
However, since map elements only have shape and semantic information, accurate association between 2D observations and 3D HD-map is still difficult. 
The repeated structures, missed detections and false detections make data association highly ambiguous, thus challenging HD-MRL methods. 
To this end, Wang \textit{et al.} \cite{Wang2021ICRA} proposed a robust data association method considering local structural consistency, global pattern consistency and temporal consistency, and introduced a sliding-window factor graph optimization framework to fuse association and odometry measurements. 
In \cite{Lu2017IV,Pauls2020IROS}, authors adopted the Semantic Chamfer Matching (SCM) algorithm to perform 2D-3D data association. 
In these works, SCM is utilized as a general cost model $d(\cdot,\cdot)$ for different map elements to measure the re-projection errors between observed and projected map elements. 
As a further improvement, TM${}^3$Loc \cite{TM3Loc} derives an analytical derivation of SCM cost with respect to the 6 DoF pose on $\mathfrak{se}(3)$ to ensure efficient optimization, avoiding the inaccuracy due to any prior assumption of the element shapes. 
With the efforts on simplified model of map elements, these works can perform HD-MRL in real-time and achieve reasonable performances for long-term localization.

Even with tremendous efforts in traditional HD-MRL methods, the accurate explicit map element matching and pose estimation are still hard problems.
In last two years, some DL-based HD-MRL methods have been developed to perform data association in DL feature domains. 
OrienterNet \cite{sarlin2023orienternet} estimates 3 DoF pose in a 2D map, OpenStreetMap (OSM), by proposed neural map matching. 
U-BEV \cite{Camiletto2023UBEVHB} leverages a U-Net \cite{UNet} inspired network to convert surround images and Standard Definition Map (SD Map) into neural representations, \textit{i.e.}, deep CNN features so that 3 DoF pose can be estimated by template matching
But their localization accuracy is only limited to sub-meter level due to the low precision of used map. 
BEV-Locator \cite{bevlocator} formulates this problem as an E2E learning scheme and proposes a Transformer-based architecture \cite{transformer} to address the key challenge of the cross-modality matching for HD-MRL. It encodes the map elements from discrete points into structured vectors, and conduct interaction between images and HD map on BEV space. EgoVM \cite{he2023egovm} extracts BEV features and map embeddings by Transformer Decoder \cite{transformer}, and then map embeddings are compared with interpolated map features by candidate poses to calculate their similarities in template matching framework, so as to estimate the optimal pose offset. These two methods with HD map achieves centimeter-level localization accuracy, but BEV-Locator \cite{bevlocator} and EgoVM \cite{he2023egovm} needs multiple view images as input, learning-based HD-MRL problem still be unsolved.

{
\vspace{4pt}
\setlength{\parindent}{0cm}
\textbf{Summary:} HD map should be the most compact and lightwight scene map, so it is preferred by AD researches. Although it needs heavy construction and maintenance cost, the recently proposed crowdsourced map building \cite{crowdsourcedmapping} and updating \cite{GCDL} strategies may help to reduce the maintenance cost, making HD map more widely acceptable.
MRL in HD map is still a very hard problem because precise data association between visual observations and semantic elements in HD map is theoretically difficult. Traditional methods seek aids from multiple sensor fusion or online odometry while recent DL-based methods leverage advanced representation ability of DNN models. Based on the localization performance and required sensor setup, we believe DL-based method is a more promising way for HD-MRL. 
}

\section{Learnt Implicit Map}
\label{sec:nn}

\begin{figure*}[!t]
    \centering
    \includegraphics[width=0.97\linewidth]{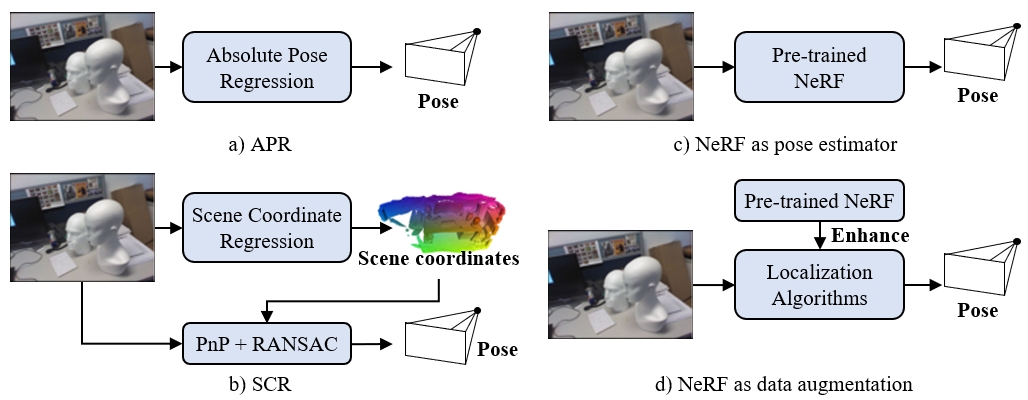}
    \caption{Diagrams of MRL method with implicit map, such as a) APR methods, b) SCR methods, and c) and d) NeRF-based methods.}
    \label{fig:nn}
\end{figure*}

In the current DL age, scholars begin to rethink to represent scene map in a more implicit way. Some recently proposed works encode scene map into neural networks so that the network can achieve amazing things, such as directly recover pose of images (called Absolute Pose Regression, APR), estimate 3D coordinates of each pixel in images (called Scene Coordinate Regression, SCR), or render geometry structure and appearance of scene (called Neural Radiance Field, NeRF), as shown in Fig. \ref{fig:nn}. Benefit from high-performance computation device, matured DL technology, and sufficient training samples, such implicit map may be a potential choice for MRL methods (called NN-MRL) in the near future.

\subsection{Absolute Pose Regression}
\label{sec:APE}
APR solution was first proposed by PoseNet \cite{posenet} to directly regress the 6 DoF pose $\textbf{x}^{\mathcal{G}}_v$ of the camera with regard to global frame $\mathcal{G}$, given a \textit{query} image $\mathcal{I}_q$. The original PoseNet \cite{posenet} is composed by a feature encoder that is a GoogLeNet backbone \cite{googlenet}, and a pose regressor that is a MLP head.
Although early APR methods achieve less accuracy than the VL-MRL methods, they enable pose estimation using a single forward pass and remove the need of saving scene map. The scene map is represented in a totally implicit and very compact formulation, \textit{i.e.}, parameters of DNN $\mbox{APR}(\cdot)$. Thus, APR method can be simply formatted as:

\begin{equation}
    \label{equ:apr1}
    \hat{\textbf{x}}^\mathcal{G}_{q} = \mbox{APR}(\mathcal{I}_q)
\end{equation}

{
\vspace{4pt}
\setlength{\parindent}{0cm}
\textbf{Single Scene APR (SS-APR):} 
In order to improve the localization accuracy, contemporary APR methods achieve further improvements by developing different CNN backbones \cite{Shavit2020CoRR,Tayyab2017IROS,hourglassPN} or MLP heads \cite{Tayyab2017IROS}, modelling localization uncertainty \cite{BayesianPN}, or introducing Long-Short-Term-Memory (LSTM) layers \cite{LSTMPN} in the networks. Since the scale of orientation (in degrees) and position (in meters) are quite different across scenes, APR methods need to use weight to balance these two factors. In original PoseNet \cite{posenet}, the scale between orientation error and translation error varies in different scenes, \textit{e.g.}, 120 to 750 for indoor scenes while 250 to 2000 for outdoor scenes, which is tricky. Kendall \textit{et al.} \cite{PoseNet2} proposed a new loss function with learnable weight of orientation error and translation error so that manual fine-tunning can be avoided, which is popularly used in later works. In \cite{mapnet}, authors improved by using logarithm of a unit quaternion instead of traditional unit quaternion as a more advanced orientation representation, and introduced geometric constraints from additional sensors, such as VO and GPS, as new loss terms in network training to improve the performance. In \cite{Saha2018ImprovedVR}, SS-APR problem is split into two step that first predicts the most relevant anchor point uniformly defined in the scene and then regress the relative offsets with respect to it. Such a scheme simplifies the APR problem and significantly improve the accuracy. For higher localization accuracy, Wang \textit{et al.} \cite{Wang2019AtLocAG} applied self-attention to the CNN features to guide the pose regression while CaTiLoc \cite{catiloc} exploits attention mechanism on full receptive field by Transformer \cite{transformer}. Xue \textit{et al.} \cite{LSG} incorporated L-RPR and APR in a single DNN and jointly trained the network so that the uncertainty of APR can be alleviated by augmenting the observation based on the co-visibility from VO estimation. In testing stage, Brahmbhatt \textit{et al.} \cite{mapnet} and Xue \textit{et al.} \cite{LSG} performed a standard Pose Graph Optimization (PGO) by optimizing only the nodes (APR estimation) with edges (VO or RPR estimation) fixed.
}

Although many modifications are proposed to the architecture and training method originally formulated by PoseNet \cite{posenet}, the main paradigm remained the same that employs 1) a CNN backbone to generate features, which is considered to contain pose-related information, and 2) a regressor to estimate absolute pose based on extracted features. The backbone and regressor are E2E trained for each scene in the SS-APR methods mentioned above.

{
\vspace{4pt}
\setlength{\parindent}{0cm}
\textbf{Multiple Scene APR (MS-APR):} 
MS-APR methods are later developed aiming to extend a single trained APR model on multiple scenes. Blanton \textit{et al.} \cite{MSPN} first proposed multi-scene PoseNet to first classifies the scene where the \textit{query} image is taken, and then retrieve a scene-specific regressor corresponding to this scene for pose regression. They train a set of regressors, one per scene, for APR with a set of scene-specific parameterized losses. MS-Transformer \cite{MS-Transformer1,MS-Transformer2} then learn multi-scene APR with a single unified Transformers-based network \cite{transformer}, where scene-specific queries (scene encodings) are used in Transformer \cite{transformer}, so they can encode many scenes in parallel and achieve MS-APR. During pose regression, both CNN features and scenes encodings are considered.
}

{
\vspace{4pt}
\setlength{\parindent}{0cm}
\textbf{Summary:}
Although APR algorithms have been developed for almost 10 years, there is no evidence that they can scale to large scale scene. For example, on Cambridge Landmarks dataset \cite{posenet} that consists of six medium-sized scenes (about 900-5500 $\mbox{m}^2$) in an urban environment, the SOTA MS-Transformer \cite{MS-Transformer2} only achieves an average localization accuracy of 0.98 m and 3.10${}^\circ$. APR methods can serve as an initial or coarse pose estimation in some real-time application, and its $O(1)$ computational complexity is also remarkable.
}

\subsection{Scene Coordinate Regression}
\label{sec:SCR}

SCR methods follow the PnP-based solution of MRL methods but in a E2E learning-based way. The scene map $\mathcal{M}$ is also represented as network parameters of DNN $\mbox{SCR}(\cdot)$. Specifically, SCR methods directly learn to predict 3D scene coordinates $\textbf{P}_i$ of each 2D pixel $\textbf{p}^{q}_i$ in the $\mathcal{I}_q$ and thus \textit{query} pose can be solved as a PnP problem based on estimated 2D-3D correspondences:
\begin{equation}
    \begin{split}
        \label{equ:scr1}
        \{\hat{\textbf{P}}_i\} &= \mbox{SCR}(\mathcal{I}_q) \\
        \hat{\textbf{x}}^\mathcal{G}_q &= \mbox{SolvePnP} (\{\textbf{p}^{q}_i, \hat{\textbf{P}}_i\}_i)
    \end{split}
\end{equation}
where $\{\textbf{p}^{q}_i, \hat{\textbf{P}}_i\}_i$ is a pair of image pixel and estimated scene coordinate.

In this field, early approaches \cite{SCRForests,Abner2014CVPR,Julien2015CVPR} used regression forest models and they required RGB-D images as input. More recent works \cite{DSAC,DSAC2,HSCNet,few-shot-SRC,DSM,KFNet,SANet,Less_is_More,Huang2021VSNetVW} have instead applied DNN-based models fed by RGB images, which greatly improves the usability of the SCR solution.

{
\vspace{4pt}
\setlength{\parindent}{0cm}
\textbf{Scene-specific SCR (SS-SCR):} 
SCR method is firstly used to estimate dense 3D coordinates for $\mathcal{I}_q$ in a previously seen scene, and it need to be specifically trained for each scene, called scene-specific methods. To reduce the localization uncertainty caused by imperfect network estimation, the PnP solver is always integrated in a RANSAC framework so that an estimation with the most inliers will be selected. However, vanilla RANSAC is non-differentiable, blocking E2E training for SCR models. So, Brachmann \textit{et al.} \cite{DSAC,DSAC2,ESAC,Less_is_More} proposed differentiable RANSAC strategies for SCR so that pose can be E2E estimated. For instance, in DSAC \cite{DSAC,DSAC2}, Brachmann \textit{et al.} scored estimated scene coordinates (can be seen as 2D-3D correspondences) using a CNN and selected final pose by probabilistic selection while in \cite{ESAC}, authors proposed a novel RANSAC framework called Expert Sample Consensus (ESAC) that many SCR networks are regard as a set of ``experts'' and a gating network is trained to measure the relevance of each expert regarding $\mathcal{I}_q$. 
}

For effectively encode scene map in SCR model, some works turn to learn region-wise label for each pixel. HSCNet \cite{HSCNet} predicts scene coordinates in a coarse-to-fine manner that it first classifies pixels into corresponding sub-regions and predict scene coordinates to reduce prediction ambiguity. Such a joint classification and regression strategy is then well studied \cite{Budvytis2019LargeSJ,few-shot-SRC,Huang2021VSNetVW,Wang2023HSCNetHS}. This way is concluded to alleviate the dependency on dense ground-truth scene coordinates in SCR model training \cite{Wang2023HSCNetHS}. Specifically, HSCNet++ \cite{Wang2023HSCNetHS} proposes a pseudo-labelling method, where ground-truth labels at each pixel location are propagated to a fixed spatial neighbourhood. However, SS-SCR trains models per scene and commonly requires pixel-wise ground-truth scene coordinates. The heavy training costs of SS-SCR block its developments.

{
\vspace{4pt}
\setlength{\parindent}{0cm}
\textbf{Scene-agnostic SCR (SA-SCR):} To boost the generalization ability and scalability of SCR methods, SA-SCR methods are proposed \cite{SANet,DSM,Revaud2023SACRegSC}. They can regress dense scene coordinates based on some given \textit{reference} views in unseen scenes.
SANet \cite{SANet} constructs a scene pyramid by extracting MS feature map from \textit{reference} images, then predicts scene coordinates for the \textit{query} image with the assistance of the scene pyramid.
In \cite{DSM}, authors proposed a Dense Scene Matching (DSM) module. The DSM module predicts scene coordinates by cost volume constructed between \textit{query} image and \textit{reference} image with its scene coordinates.
Following this way, the Transformer-based SAReg \cite{Revaud2023SACRegSC} takes a variable number of images (including \textit{query} images and retrieved \textit{reference} images) and sparse 2D-3D annotations from SfM database \cite{colmap} as inputs.
}

More implicitly, NeuMap \cite{neumap} encodes the scene as sparse scene-specific latent codes and estimates scene coordinates with the help of the resulting latent codes. When localizing in a new scene, NeuMap \cite{neumap} only need to generate new latent code for the new scene without re-training other parts of network. By pruning redundant latent code, NeuMap \cite{neumap} can achieves similar performance as DSAC++\cite{Less_is_More} with a 100 to 1000 times smaller scene map size. 
Instead of densely estimate scene coordinates, NeuMap \cite{neumap} and D2S \cite{Bui2023D2SRL} estimate scene coordinates for sparse 2D key points in \textit{query} images, making them effective. D2S \cite{Bui2023D2SRL} also supports updating by self-supervised learning procedure with new observations without requiring camera poses, intrinsic parameters, and ground-truth scene coordinates, which greatly improves the scalability and generalization ability of SCR.

{
\vspace{4pt}
\setlength{\parindent}{0cm}
\textbf{Summary:}
SCR methods do not explicitly rely on local feature extraction and matching, and they are able to provide dense correspondences between \textit{query} image and scene map.
They also do not require storing the 3D models and descriptors, as scene map is implicitly encoded in a learnable DNN.
The localization accuracy of SCR is impressively great in small- and medium-scale scenarios \cite{7scenes,posenet}, but the expensive training costs block its usage in unbounded scene.
}

\subsection{Neural Radiance Field}
\label{sec:NeRF}

NeRF implicitly represents a scene as a ``radiance field'': a volumetric density that models the geometry of the scene, and a view-dependent color that models the appearance of occupied regions in the scene \cite{NeRF}. 
A Multiple Layer Perception (MLP) is usually utilized to takes the 3D position of a point $\textbf{P}$ and the unit-norm viewing direction of that point $\textbf{v}$ as inputs, and estimate the density $\sigma$ and color $c$ of the point, $(\sigma,c) \leftarrow \mbox{NeRF}(\textbf{P},\textbf{v})$. Scene map is implicitly represented by the parameters $\Theta$ of $\mbox{NeRF}(\cdot,\cdot)$. Given some RGB images $\{\mathcal{I}_t\}$ with known camera poses $\{\textbf{x}^\mathcal{G}_t\}$, NeRF is trained by minimizing a photometric loss $\mathcal{L}$ \cite{NeRF}:
\begin{equation}
    \label{equ:nerf1}
    \hat{\Theta} = \mathop{\arg\min}\limits_{\Theta} \mathcal{L} (\{\mathcal{I}_t\}, \{\textbf{x}_t\})
\end{equation}

NeRF learns to synthesize novel views associated with any camera poses, enabling virtual view rendering and scene geometry recovery, thus have been applied in MRL in recent years. Some works directly uses pre-trained NeRF to estimate pose while other works utilize NeRF as a novel data augmentation method to enrich the training data.

{
\vspace{4pt}
\setlength{\parindent}{0cm}
\textbf{NeRF as pose estimator:} As the first framework to estimate pose using NeRF, iNeRF \cite{iNeRF} directly optimizes 6 DoF poses with freezed pre-trained network parameters $\Theta$ by minimizing the photometric error between rendered pixels and observed pixels, which is an inverse problem of original NeRF \cite{NeRF}:
}
\begin{equation}
    \label{equ:nerf2}
    \hat{\textbf{x}}_q = \mathop{\arg\min}\limits_{\textbf{x}_q \in SE(3)} \mathcal{L} (\mathcal{I}_q, \textbf{x}_q~|~\Theta~\mbox{freezed})
\end{equation}

In a very similar way, NeRF-- \cite{nerfmm} simplifies the training process of NeRF on forward-facing scenes by removing the need of camera pose and intrinsics. Later, SiNeRF \cite{SiNeRF} improves by leverage sinusoidal activations for radiance mapping and proposing a novel mixed region sampling to select ray batch efficiently. These joint pose estimation and NeRF training methods then motivate pose-free NeRF \cite{Meng2021GNeRFGN,NoPe-NeRF}.

Many methods aiming at pose estimation only are also developed.
In LATITUDE \cite{LATITUDE}, authors trained a pose regressor through images generated from pre-trained NeRFs, which provides an initial pose guess for hierarchical localization without the requirement of saving reference images. In fine pose optimization stage, LATITUDE \cite{LATITUDE} minimizes the photometric error between the observed image and rendered image by optimizing the pose on the tangent plane. 
In Loc-NeRF \cite{LocNeRF}, a pre-trained NeRF model is integrated into a Monte Carlo localization to measure weights to particles by image similarity between \textit{query} image and rendered image. 
In NeRF-Loc \cite{liu2023nerfloc}, a NeRF conditioned on \textit{reference} images takes 3D point as input and generates corresponding 3D descriptors so that 2D-3D matches can be obtained by matching. Precise 6 DoF pose is then estimated by a PnP solver.

\begin{figure}[!h]
    \centering
    \includegraphics[width=0.90\linewidth]{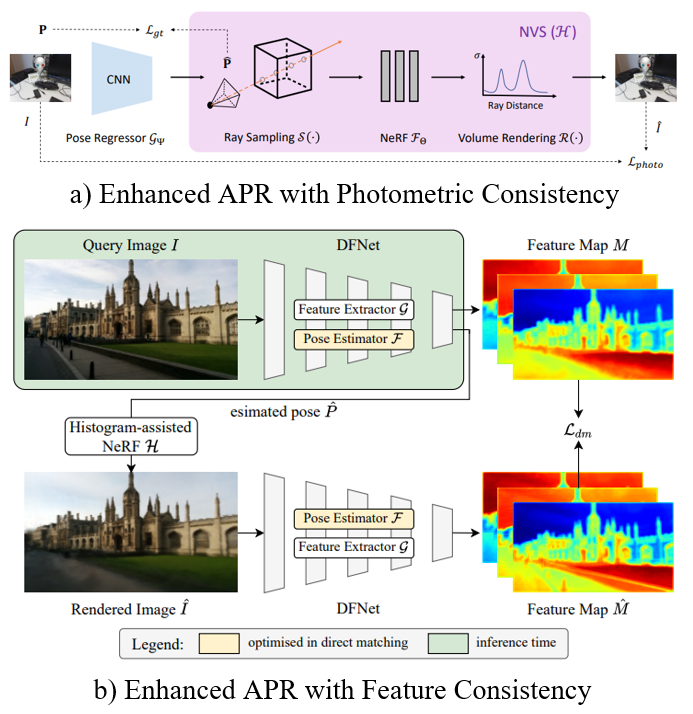}
    \caption{Instances of APR method enhanced by NeRF, a) Direct-PoseNet \cite{DPN} using photometric consistency between \textit{query} image and rendered image, while b) DFNet \cite{DFNet} using feature consistency.}
    \label{fig:APR+NeRF}
\end{figure}

{
\vspace{4pt}
\setlength{\parindent}{0cm}
\textbf{NeRF as data augmentation:} As a more straightforward way to apply NeRF in MRL, NeRF can be used to render virtual images at unseen viewpoints so that the overfitting problem of APR methods can be effectively solved and thus improves performances of APR \cite{Moreau2021LENSLE,DPN,DFNet}. LENS \cite{Moreau2021LENSLE} applies pre-trained NeRF-W \cite{NeRF-W} as an offline data augmentation pipeline to enhance APR model. As an online augmentation way, Direct-PoseNet \cite{DPN} minimizes the photometric errors between \textit{query} image and virtual image rendered at estimated pose as shown in Fig. \ref{fig:APR+NeRF} a) while DFNet \cite{DFNet} uses feature map errors \cite{DFNet} as shown in Fig. \ref{fig:APR+NeRF} b), making APR method more precise and scalable. Enhanced by NeRF, APR methods can achieve significantly improved localization accuracy.
}

{
\vspace{4pt}
\setlength{\parindent}{0cm}
\textbf{Summary:}
As a newly emerged technique, NeRF has two favourable characterises: 1) the implicit representation of scene; 2) the ability to render synthesis image at novel viewpoint. These two characterises present great potential in MRL researches. But we have to say the utilization of NeRF in MRL is still an opening question to be explored.
}

\section{Benchmarking Monocular Re-Localization}
\label{sec:benchmark}

In this section, the common evaluation metrics and datasets are introduced, and then we provide the performances of some SOTA MRL methods with various type of scene map on these benchmark. 
Looking back to all kind of MRL methods mentioned above, some of MRL methods retrieve \textit{reference} image as \textit{query} pose approximation while others perform precise pose estimation for \textit{query} image. We name these two kind of MRL methods as Coarse Pose Approximation (CPA) and Fine Pose Estimation (FPE), respectively, as listed in Tab. \ref{tab:PA-PE}. These two kinds of MRL are generally tested using different evaluation metric. We provide introduction and evaluation respectively.

\begin{table}[!h]
    \centering
    \renewcommand{\arraystretch}{1.5}
    \caption{The categorisation of various MRL methods based on the function and their evaluation metrics.}
    {
        \begin{tabularx}{\linewidth}{Xll}
            \hline
            \hline
            Function & Method & Evaluation Metric\\
            \hline
            \makecell[l]{Coarse Pose\\ Approximation\\ (CPA)} & \makecell[l]{VPR (Sec. \ref{sec:VPR}),\\ I2P-VPR (Sec. \ref{sec:learning-based i2p})} & \makecell[l]{Precision (P), Recall (R),\\ R@K, maxP@R=1} \\
            \hline
            \makecell[l]{Fine Pose\\ Estimation\\ (FPE)} & \makecell[l]{RPR (Sec. \ref{sec:RPR}),\\ VL-MRL (Sec. \ref{sec:visual}),\\ G-I2P (Sec. \ref{sec:geo-based i2p}),\\  I2P-RPR (Sec. \ref{sec:learning-based i2p}),\\ I2P-MLoc (Sec. \ref{sec:learning-based i2p}),\\ HD-MRL (Sec. \ref{sec:hdmap}),\\ NN-MRL (Sec. \ref{sec:nn})} & \makecell[l]{Position error $\mbox{err}_{pos}$,\\ Orientation error $\mbox{err}_{rot}$} \\
            \hline
            \hline
        \end{tabularx}
    }
    \label{tab:PA-PE}
\end{table}

\begin{table*}[!h]
    \centering
    \renewcommand{\arraystretch}{1.3}
    \caption{Some datasets used as evaluation benchmarks in this paper.}
    {
    \resizebox{\linewidth}{!}{
        \begin{tabular}{lll}
            \hline
            \hline
            Datasets & Description & Evaluated Methods \\
            \hline
            GP \cite{GardenPoint2} & campus, 3 traverses with the same trajectory (2 at daytime while 1 at nighttime) & \multirow{9}{*}{VPR (Sec. \ref{sec:VPR})} \\
            SPEDTest \cite{SPED} & collected from all over the world, weather, seasonal and illumination changes & \\
            Nordland \cite{seqslam}& natural and outdoor scenes, seasonal changes across different traverses &\\
            Pittsburgh \cite{pittsburgh} & collected from Google Street View, significant viewpoint changes & \\
            MSLS \cite{MSLS} & collected by ground vehicles, viewpoint and illumination changes & \\
            Oxford \cite{fabmap} & including CityCentre and NewCollege, campus, fish-eye binocular images & \\
            St. Lucia \cite{RatSLAM} & suburb, viewpoint and day-night appearance changes, dynamic objects & \\
            17-Places \cite{17places} & indoor scenes, viewpoint changes & \\
            Baidu Mail \cite{BaiduMail} & indoor scenes, reflectance and repetitive structures & \\
            \hline
            Aachen Day-Night \cite{benchmark6dof} & \makecell[l]{\textit{query} images taken at daytime and \textit{reference} images taken at nighttime, significant\\ illumination and viewpoint changes} & \multirow{2}{*}{VL-MRL (Sec. \ref{sec:visual})} \\
            InLoc \cite{InLoc} & indoor scenes, large viewpoint and illumination changes, moving furniture, occlusions & \\
            \hline
            KAIST \cite{KAIST} & highly complex urban for AD & HD-MRL (Sec. \ref{sec:hdmap}) \\
            \hline
            7Scenes \cite{posenet} & RGB-D images with GT poses are provided, 7 different indoor scenes & \multirow{4}{*}{ALL} \\
            \makecell[l]{Cambridge \\Landmarks \cite{posenet}} & 6 medium-scale outdoor scenes &  \\
            KITTI \cite{KITTI} & \makecell[l]{collected by ground vehicles, 11 of 22 sequences have GT trajectories, both images\\ and synchronized LiDAR scans are provided} & \\
            \hline
            \hline
        \end{tabular}
    }
    }
    \label{tab:dataset}
\end{table*}

\subsection{Evaluation metrics}
\label{sec:metric}

\subsubsection{MRL~~for~~CPA}

Those MRL methods for CPA utilize evaluation metric as image retrieval task.
A retrieved \textit{reference} image/point cloud clip is defined as true positive if it was geometrically near the \textit{query} images. The ground-truth is generally labeled by geometric distance based on GPS signals \cite{Li2023I2PRecRI} or manual annotation \cite{yue1,yue2}. With the ground-truth, we can measure Precision (P) and Recall (R) metric, where precision denotes the ratio between true positives and all the retrieved \textit{references} and recall is the ratio between the number of true positives and all the potentially matched \textit{query}-\textit{references} existed in this scene.

In some MRL application need to retrieve comprehensively, we measure the ratio of retrieved true positives over top-K candidates, R@K.
In some applications that a mismatch will make the whole system dramatically fails, \textit{e.g.}, loop closure detection in SLAM \cite{LCD2022TITS}, we turn to measure the maximum recall metric with 100\% precision, maxP@R=1. The higher these two metrics, the better a MRL method performs.

\subsubsection{MRL~~for~~FPE}

Direct measurement about estimated poses is applied in those MRL methods aiming at FPE. The common evaluation metrics are position error $\mbox{err}_{pos}$ for translation $\textbf{t}^\mathcal{G}_v$ (in distance) and orientation error $\mbox{err}_{rot}$ for rotation $\textbf{R}^\mathcal{G}_v$ (in degree) between estimated results and ground-truth poses. The lower these two errors, the better a MRL method performs. In common scenes that vehicles move in 6 DoF, the position error and orientation error are both 3 DoF. But in AD scene for ground vehicles, we measure 2 DoF position error for lateral and longitudinal position, and 1 DoF orientation error for heading angle.

Additionally, the size of represented scene map $\mathcal{M}$ and the running efficiency of MRL methods should also be considered.

\subsection{Dataset}
\label{sec:dataset}
There are numerous datasets available for the evaluation of MRL methods and it is difficult to give a complete overview. Here we only introduce some datasets that we used to evaluate MRL methods in this survey, as listed in Tab. \ref{tab:dataset}.

\subsubsection{MRL~~for~~CPA}
The dataset used to evaluate VPR is required to include \textit{query} and \textit{reference} images that can be paired by geometric distance. So the dataset should have traversed the same scene multiple times, or the trajectory of the dataset should be ``looped'', such as \textbf{Gardens Point (GP) dataset} \cite{GardenPoint2}, \textbf{SPEDTest dataset} \cite{SPED}, \textbf{Nordland dataset} \cite{nordland}, \textbf{Pittsburgh dataset} \cite{pittsburgh}, \textbf{Mapillary Street-Level Sequences (MSLS) dataset} \cite{MSLS}, \textbf{Oxford CityCentre and NewCollege dataset} \cite{fabmap}, \textbf{St. Lucia dataset} \cite{RatSLAM}, \textbf{17-Places dataset} \cite{17places}, \textbf{Baidu Mail dataset} \cite{BaiduMail}, Tokyo 24/7 dataset \cite{tokyo247}, Synthia dataset \cite{SYNTHIA}, \textit{etc.}

To evaluate MRL for CPA (including both VPR and I2P-VPR), it requires the dataset to have both images and LiDAR sweeps. Since I2P-VPR is a newly emerged research, there is very few dataset proposed for the evaluation in this field. Researches in this field often use \textbf{KITTI dataset} \cite{KITTI}. Based on the given pose of each image in KITTI \cite{KITTI}, ground-truth matches between \textit{query} and \textit{reference} image/point cloud can be labelled, and both VPR and I2P-VPR can be evaluated since synchronized LiDAR scans are also provided. We generally offline build LiDAR point cloud map using LiDAR SLAM to generate \textit{reference} point cloud clip.

\subsubsection{MRL~~for~~FPE}

Some datasets provide ground-truth pose for each image so that we can evaluate the MRL methods aiming at FPE. 
For example, the outdoor \textbf{Aachen Day-Night dataset} \cite{benchmark6dof} and the indoor \textbf{InLoc dataset} \cite{InLoc} are usually utilized to evaluate VL-MRL methods. \textbf{KAIST Urban dataset} \cite{KAIST} provides LiDAR data and stereo image with various sensors targeting a highly complex urban environment. Some works \cite{TM3Loc,Liao2019CoarseToFineVL} build HD map for KAIST \cite{KAIST} so that it can be used to evaluate HD-MRL methods. \textbf{KITTI dataset} \cite{KITTI} provides ground-truth pose for each image, so it can also be used to evaluate MRL for FPE. Since multiple sensor data is provided, it can support evaluation for all kind of MRL. The indoor \textbf{7Scenes dataset} \cite{posenet} and outdoor \textbf{Cambridge Landmarks dataset} \cite{posenet} provides ground-truth camera pose and 3D scene model, so they also support the evaluation of almost all the kinds of MRL methods. We compare most MRL methods on these two dataset to analyze the strengths and drawbacks of each kind of MRL methods. Except to these datasets, CMU Seasons and Extended CMU Seasons \cite{benchmark6dof}, RobotCar Seasons \cite{benchmark6dof}, ETH-Microsoft Dataset \cite{eth_ms_visloc_2021} are also widely used in evaluation, Map-free Localization dataset \cite{arnold2022mapfree} provides a specialised benchmark for evaluating RPR methods when only one \textit{reference} image is included in the scene map.

\begin{table*}[!h]
    \renewcommand{\arraystretch}{1.3}
    \centering
    \caption{Recall over top-K candidates of different GF-VPR techniques on popular benchmarks.}
    \begin{threeparttable}
        \resizebox{1.0\linewidth}{!}{
            \begin{tabular}{l|c|ccc|ccc|ccc|ccc}
                \hline
                \hline
                \multirow{2}{*}{Method}                   & \multirow{2}{*}{dim} & \multicolumn{3}{c|}{Pitts250k-test \cite{pittsburgh}} & \multicolumn{3}{c|}{MSLS-val \cite{MSLS}} & \multicolumn{3}{c|}{SPED \cite{SPED}} & \multicolumn{3}{c}{Nordland \cite{nordland}}                                                         \\
                                                          &                      & R@1                                                   & R@5                                       & R@10                                  & R@1                                          & R@5  & R@10 & R@1  & R@5  & R@10 & R@1  & R@5  & R@10 \\
                \hline
                GeM \cite{GeM} ${}^\dagger$               & 2048                 & 72.3                                                  & 87.2                                      & 91.4                                  & 65.1                                         & 76.8 & 81.4 & 55.0 & 70.2 & 76.1 & 7.4  & 13.5 & 16.6 \\
                NetVLAD \cite{netvlad} ${}^\dagger$       & 32768                & 86.0                                                  & 93.2                                      & 95.1                                  & 59.5                                         & 70.4 & 74.7 & 71.0 & 87.1 & 90.4 & 4.1  & 6.6  & 8.2  \\
                \hline
                GeM \cite{GeM} ${}^\star$                 & 2048                 & 82.9                                                  & 92.1                                      & 94.3                                  & 76.5                                         & 85.7 & 88.2 & 64.6 & 79.4 & 83.5 & 20.8 & 33.3 & 40.0 \\
                NetVLAD \cite{netvlad} ${}^\star$         & 32768                & 90.5                                                  & 96.2                                      & 97.4                                  & 82.6                                         & 89.6 & 92.0 & 78.7 & 88.3 & 91.4 & 32.6 & 47.1 & 53.3 \\
                CosPlace \cite{cosplace} ${}^\star$       & 2048                 & 91.5                                                  & 96.9                                      & 97.9                                  & 84.5                                         & 90.1 & 91.8 & 75.3 & 85.9 & 88.6 & 34.4 & 49.9 & 56.5 \\
                MixVPR \cite{mixvpr} ${}^\star$           & 4096                 & 94.6                                                  & 98.3                                      & 99.0                                  & 88.0                                         & 92.7 & 94.6 & 85.2 & 92.1 & 94.6 & 58.4 & 74.6 & 80.0 \\
                \hline
                \hline
            \end{tabular}
        }
        \resizebox{1.0\linewidth}{!}{
            \begin{tabular}{l|cc|cc|cc|cc|cc|cc}
                \multirow{2}{*}{Method}                    & \multicolumn{2}{c|}{Baidu Mall \cite{BaiduMail}} & \multicolumn{2}{c|}{Gardens Point \cite{GardenPoint2}} & \multicolumn{2}{c|}{17 Places \cite{17places}} & \multicolumn{2}{c|}{Pitts-30k \cite{pittsburgh}} & \multicolumn{2}{c|}{St. Lucia \cite{RatSLAM}} & \multicolumn{2}{c}{Oxford \cite{oxford}}                                                 \\
                                                           & R@1                                              & R@5                                                    & R@1                                            & R@5                                              & R@1                                           & R@5                                      & R@1   & R@5   & R@1   & R@5   & R@1   & R@5   \\
                \hline
                NetVLAD \cite{netvlad}                     & 53.10                                            & 70.51                                                  & 58.50                                          & 85.00                                            & 61.58                                         & 77.83                                    & 86.08 & 92.66 & 57.92 & 72.95 & 52.88 & 74.87 \\
                CosPlace \cite{cosplace}                   & 41.62                                            & 55.02                                                  & 74.00                                          & 94.50                                            & 61.08                                         & 76.11                                    & 90.45 & 95.66 & 99.59 & 99.93 & 91.10 & 99.48 \\
                MixVPR \cite{mixvpr}                       & 64.44                                            & 80.28                                                  & 91.50                                          & 96.00                                            & 63.79                                         & 78.82                                    & 91.52 & 95.47 & 99.66 & 100   & 90.05 & 98.95 \\
                AnyLoc-VLAD-DINOv2 \cite{keetha2023anyloc} & 75.22                                            & 87.57                                                  & 95.50                                          & 99.50                                            & 65.02                                         & 80.54                                    & 87.66 & 94.69 & 96.17 & 98.84 & 98.95 & 100   \\
                \hline \hline
            \end{tabular}
        }
        \begin{tablenotes}
            \item[1] The results are cited from MixVPR \cite{mixvpr} and AnyLoc \cite{keetha2023anyloc}.
            \item[$\dagger$] The results reported by the original publications.
            \item[$\star$] The variants are trained on GSVcities dataset \cite{GSVCities} using the same backbone network (ResNet-50 \cite{resnet}) with MixVPR \cite{mixvpr}.
        \end{tablenotes}
    \end{threeparttable}
    \label{tab:vprbench1}
\end{table*}

\begin{table*}[!t]
    \renewcommand{\arraystretch}{1.3}
    \tabcolsep=5pt
    \centering
    \begin{threeparttable}
        \caption{Maximum recall at 100$\%$ precision of different VPR methods on public benchmarks.}
        \label{tab:vprbench2}
        \begin{tabular}{l c | c c c c c c c c }
            \hline
            \hline
            \multirow{2}{*}{Methodology}                 & \multirow{2}{*}{From} & \multicolumn{2}{c}{Oxford \cite{fabmap}} & \multicolumn{3}{c}{KITTI \cite{KITTI}} & Malaga \cite{malaga} & \multicolumn{2}{c}{St. Lucia \cite{RatSLAM}}                                                         \\
                                                         &                       & CityCentre                               & NewCollege                             & \#00                 & \#05                                         & \#06   & Parking 6L & 100909 (12:10) & 180809 (15:45) \\
            \hline
            \multicolumn{10}{l}{\textbf{Image Global Feature-based Visual Place Recognition}}                                                                                                                                                                                                      \\
            \hline
            NetVLAD \cite{netvlad}                       & TPAMI'18              & 0.7148                                   & 0.4369                                 & 0.9315               & 0.9108                                       & 0.9632 & 0.3236     & 0.8046         & 0.7967         \\
            ResNet50-AP-GeM \cite{GeM}                   & TPAMI'19              & 0.9056                                   & 0.8485                                 & 0.9124               & 0.9470                                       & 0.9779 & 0.6011     & 0.7926         & 0.6936         \\
            Kazmi \textit{et al.} \cite{Kazmi2019TRO}    & TRO'19                & 0.7558                                   & 0.5109                                 & 0.9039               & 0.8141                                       & 0.9739 & 0.5098     & 0.8006         & 0.7255         \\
            \hline
            \multicolumn{10}{l}{\textbf{Image Local Feature-based Visual Place Recognition}}                                                                                                                                                                                                       \\
            \hline
            FABMAP 2.0 \cite{fabmap2}                    & IJRR'11               & 0.3526                                   & 0.6038                                 & 0.8222               & 0.3712                                       & 0.6347 & n.a.       & n.a.           & n.a.           \\
            DLoopDetector \cite{dbow-tro}                & TRO'12                & 0.3059                                   & n.a.                                   & 0.7243               & 0.5197                                       & 0.8971 & 0.7475     & 0.3722         & 0.3136         \\
            Tsintotas \textit{et al.} \cite{AVWs}    & ICRA‘18               & 0.6595                                   & 0.2988                                 & 0.9318               & 0.9420                                       & 0.9669 & 0.8799     & 0.2627         & 0.1507         \\
            iBoW-LCD \cite{ibow-lcd}                     & RAL'18                & 0.8825                                   & 0.7940                                 & 0.7650               & 0.5307                                       & 0.9553 & 0.5098     & 0.7002         & 0.8750         \\
            SLCD \cite{slcd}                             & TII'21                & 0.4088                                   & 0.7529                                 & 0.9753               & 0.8972                                       & 0.9313 & n.a.       & n.a.           & n.a.           \\
            Yue \textit{et al.} \cite{yue2}              & JFR'22                & 0.9127                                   & 0.9463                                 & 0.9569               & 0.9518                                       & 0.9963 & 0.8992     & n.a.           & n.a.           \\
            \hline
            \multicolumn{10}{l}{\textbf{Image Global and Local Feature-based Visual Place Recognition}}                                                                                                                                                                                            \\
            \hline
            HTMap \cite{htmap}                           & TRO'17                & 0.7968                                   & 0.7360                                 & 0.9024               & 0.7588                                       & 0.9703 & n.a.       & n.a.           & n.a.           \\
            FILD \cite{fild}                             & IROS'19               & 0.6648                                   & 0.7674                                 & 0.9123               & 0.8515                                       & 0.9338 & 0.5609     & 0.7606         & 0.6696         \\
            FILD++ \cite{fild2}                          & JFR'22                & 0.9091                                   & n.a.                                   & 0.9492               & 0.9542                                       & 0.9816 & 0.6274     & 0.8339         & 0.8136         \\
            Liu \textit{et al.}\cite{liu}                & ICRA'21               & 0.8601                                   & 0.9121                                 & 0.9302               & 0.9253                                       & n.a.   & n.a.       & n.a.           & n.a.           \\
            \hline
            \multicolumn{10}{l}{\textbf{Sequence-based Visual Place Recognition}}                                                                                                                                                                                                                  \\
            \hline
            SeqSLAM \cite{seqslam}                       & ICRA'12               & 0.3186                                   & 0.5757                                 & 0.8758               & 0.1823                                       & 0.6068 & n.a.       & n.a.           & n.a.           \\
            Bampis \textit{et al.} \cite{Bampis2018IJRR} & IJRR'18               & 0.4963                                   & 0.8087                                 & 0.8947               & 0.8771                                       & 0.8015 & 0.3393     & 0.6093         & 0.4979         \\
            \hline
            \hline
        \end{tabular}
        \begin{tablenotes}
            \item[1] Most of the results are cited from recent papers \cite{yue2} and \cite{fild2}.
            \item[2] ``n.a.'' means the method cannot achieve 100\% precision in our implementation with default setups and the result in published papers is unavailable.
        \end{tablenotes}
    \end{threeparttable}
\end{table*}

\subsection{Performance of SOTAs}
\label{sec:sota}

Here, we provide the performance of some SOTA MRL methods from public publications as a reference to evaluate the advantages and disadvantages of different methods.

\subsubsection{MRL~~for~~CPA} MRL methods for CPA, including VPR and I2P-VPR, tend to retrieve \textit{reference} data given \textit{query} image, so they are measured and compared based on the precision and recall metrics of their retrieved results. 

{
\vspace{4pt}
\setlength{\parindent}{0cm}
\textbf{VPR: }
First, we give some performance of VPR methods in public datasets, evaluated with different metrics, \textit{e.g.}, R@K (in Tab. \ref{tab:vprbench1}) and maxP@R=1 (in Tab. \ref{tab:vprbench2}). It shows in Tab. \ref{tab:vprbench1} that the differentiable VLAD pooling layer could effectively aggregate dense local descriptors into global features, thus making NetVLAD \cite{netvlad} an excellent solution in retrieval-based GF-VPR tasks. Then, benefit from the global fusion of features, MixVPR \cite{mixvpr} achieves improved performance, espeically in challenging Nordland dataset \cite{nordland} with varying weather and seasons. More recently, pre-trained features from visual foundation model (such as DINOv2 \cite{oquab2023dinov2}) are concluded to be an excellent alternate for GF-VPR, and enable AnyLoc \cite{keetha2023anyloc} achieves impressive retrieval performance in very different scene. Although GF-VPR methods can fully retrieve \textit{reference} images, their retrieval accuracy is slightly lower than SOTA LF-VPR methods as shown in Tab. \ref{tab:vprbench2}. Adding verification with local features \cite{yue2} or jointly utilizing local feature and global feature \cite{fild2} can achieve pleased accuracy while retrieve most of \textit{reference} images for each \textit{query} image.
}

\begin{figure}[!h]
    \centering
    \includegraphics[width=0.97\linewidth]{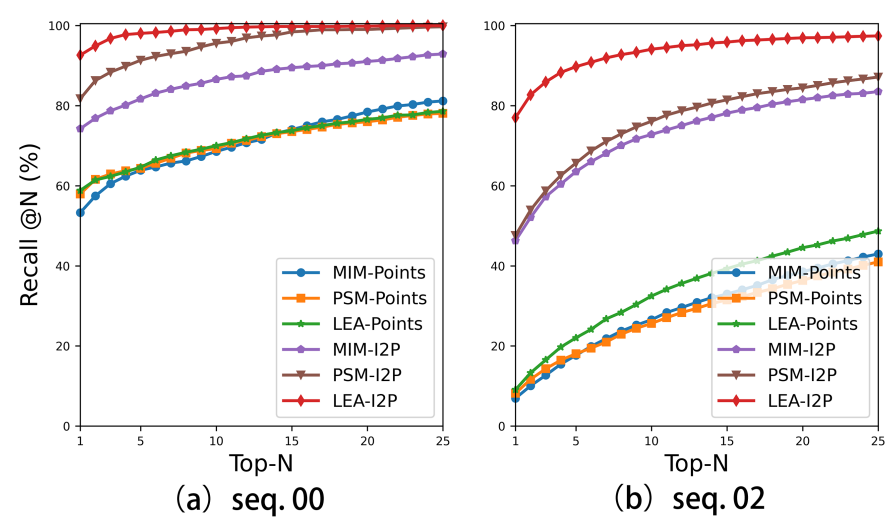}
    \caption{Recall over top-K candidates of I2P-Rec \cite{Li2023I2PRecRI} on KITTI \cite{KITTI}.}
    \label{fig:vprbench3}
\end{figure}

{
\vspace{4pt}
\setlength{\parindent}{0cm}
\textbf{I2P-VPR: }
As for cross-modal I2P-VPR between \textit{query} image and \textit{reference} point cloud clip, a SOTA performance is achieved by I2P-Rec \cite{Li2023I2PRecRI} as shown in Fig. \ref{fig:vprbench3}. With the assistance of mature monocular depth estimation, I2P-Rec \cite{Li2023I2PRecRI} transform \textit{query} image and point cloud clip from scene map to BEV images, simplifying the cross-modal retrieval problem and leading to pleased performance.
}

{
\vspace{4pt}
\setlength{\parindent}{0cm}
\textbf{Summary: }
The objective of MRL methods for CPA is to accurately and completely retrieve \textit{reference} for each \textit{query}, which requires robustly measure the similarity between \textit{query} and \textit{reference} against shifted viewpoints, changed appearance, dynamic occlusions, and so on. The utilization of pre-trained feature from computer vision community \cite{keetha2023anyloc}, geometric verification based on local features \cite{yue2}, robust representation combined local and global features \cite{fild2}, temporal enhancement \cite{Bampis2018IJRR}, and transferring into unified representation space \cite{Li2023I2PRecRI} may be helpful ways to improve performance for this field. But VPR and I2P-VPR methods directly use retrieved \textit{reference} pose as current pose, their localization accuracy is inherently limited, which will be experimental concluded in later paragraphs. So they are generally used as coarse pose guess in hierarchical localization framework. 
}

Additionally, VPR-Bench\footnote{https://github.com/MubarizZaffar/VPR-Bench} \cite{vprbench} is an open-sourced VPR evaluation framework with quantifiable viewpoint and illumination invariance, enabling fast and full quantified evaluation for VPR methods. In I2P-VPR field, such public evaluation benchmark is not proposed yet.

\subsubsection{MRL~~for~~FPE}

Several MRL methods tend to estimate fine pose. They utilize various kind of scene map and achieve different level of localization accuracy. We first provide performance of RPR, VL-MRL, I2P-MRL, HD-MRL, and NN-MRL respectively. Then, these methods are evaluated on two datasets \cite{7scenes,posenet} so that they can be fairly compared.

{
\vspace{4pt}
\setlength{\parindent}{0cm}
\textbf{RPR: }
Towards evaluation of RPR methods, a benchmark called \textit{mapfree-reloc-benchmark} has been recently proposed \footnote{https://research.nianticlabs.com/mapfree-reloc-benchmark/leaderboard}, we suggest readers to refer to the public leader board for detailed information. Up to now, the SOTA solution of RPR is solving pose by decomposing essential matrix between matched features across \textit{query} image and \textit{reference} image and then recover scale by a monocular depth estimation method \cite{DPT}. It can achieve a localization performance with 1.23m translation error and 11.1\degree ~rotation error.
}

\begin{table}[!h]
    \centering
    \renewcommand{\arraystretch}{1.3}
    \resizebox{0.5\textwidth}{!} {
    \begin{threeparttable}
    \caption{Localization success ratio of some VL-MRL methods on Aachen Day-Night Benchmark (v1.0)~\cite{benchmark6dof}.}
    \label{tab:aachen}
        \begin{tabular}{l c c}
            \hline
            \hline
            \multirow{2}{*}{Method}                  & \multicolumn{2}{c}{Localized Queries (\%, 0.25$m$,2$^\circ$/0.5$m$,5$^\circ$/1.0$m$, 10$^\circ$)}                       \\
                                                     & Day                                                                                               & Night               \\
            \hline
            \multicolumn{3}{l}{\textbf{Local Feature Evaluation on Night-time Queries}}                                                                                        \\
            \hline
            SIFT \cite{sift} 5K & - & 20.4 / 27.6 / 37.8 \\
            SIFT + CAPS~\cite{CAPS}            & -                                                                                                 & 77.6 / 86.7 / 99.0  \\
            SuperPoint~\cite{superpoint}        & -                                                                                                 & 73.5 / 79.6 / 88.8  \\
            D2-Net (SS) \cite{d2net} 1K  & - & 64.3 / 78.6 / 91.8 \\
            D2-Net (SS) \cite{d2net} 5K  & - & 77.6 / 92.9 / 100.0  \\
            D2-Net (MS) \cite{d2net} 1K  & - & 53.1 / 74.5 / 86.7 \\
            R2D2 \cite{r2d2} 1K & - & 55.1 / 70.4 / 77.6 \\
            R2D2 \cite{r2d2} 5K & - & 69.4 / 84.7 / 98.0  \\
            ASLFeat~\cite{aslfeat}              & -                                                                                                 & 77.6 / 89.8 / 100.0 \\
            SEKD (SS) \cite{sekd} 1K & - & 30.6 / 33.7 / 38.8 \\
            SEKD (SS) \cite{sekd} 5K & - & 56.1 / 69.4 / 79.6  \\
            SEKD (MS) \cite{sekd} 1K & - & 35.7 / 42.9 / 50.0 \\
    	SEKD (MS) \cite{sekd} 5K & - & 67.3 / 74.5 / 83.7  \\
    	DISK \cite{disk} 1K & - & 60.2 / 72.4 / 81.6 \\
            DISK \cite{disk} 5K &  - & 80.6 / 86.7 / 99.0  \\
    	GLFeat \cite{glfeat} 5K  & - & 79.6 / 86.7 / 100.0  \\
            ALIKE-L \cite{alike} 1K  & - & 59.2 / 73.5 / 83.7 \\
            ALIKED-N(32) \cite{aliked} 1K & - & 77.6 / 88.8 / 100.0 \\
            DualRC-Net~\cite{DRCNet}                & -                                                                                               & 79.6 / 88.8 / 100.0 \\
            SparseNCNet~\cite{SparseNCNet}           & -                                                                                                 & 76.5 / 84.7 / 98.0  \\  
            Patch2Pix \cite{patch2pix}             & -                                                                                                 & 79.6 / 87.8 / 100.0 \\    		   		
            \hline
            \multicolumn{3}{l}{\textbf{Full Localization with HLoc~\cite{hloc2}}}                                                                                              \\
            \hline
            SuperPoint~\cite{superpoint}        & 85.4 / 93.3 / 97.2                                                                                & 75.5 / 86.7 / 92.9  \\
            \makecell[l]{SuperPoint +\\~~SuperGlue~\cite{superglue}}  & 89.6 / 95.4 / 98.8                                                                                & 86.7 / 93.9 / 100.0 \\
            Patch2Pix \cite{patch2pix}               & 84.6 / 92.1 / 96.5                                                                                & 82.7 / 92.9 / 99.0  \\
            \makecell[l]{Patch2Pix\\~~(w.CAPS) \cite{patch2pix}}      & 86.7 / 93.7 / 96.7                                                                                & 85.7 / 92.9 / 99.0  \\
            \makecell[l]{Patch2Pix\\~~(w.SuperGlue) \cite{patch2pix}} & 89.2 / 95.5 / 98.5                                                                                & 87.8 / 94.9 / 100.0 \\ 
            \hline
            \hline
        \end{tabular}
        \begin{tablenotes}
            \item[1] Some results are cited from \cite{patch2pix} and we do not know the limitation of number of features. The results with limited number of features are denotes like ``SIFT 5K'' and cited from \cite{aliked,glfeat}.
        \end{tablenotes}
    \end{threeparttable}
    }
    
\end{table}

{
\vspace{4pt}
\setlength{\parindent}{0cm}
\textbf{VL-MRL: }
As for VL-MRL methods, as shown in Tab. \ref{tab:aachen} and Tab. \ref{tab:inloc}, we report the ratio of successfully localized \textit{queries} within three pre-defined tolerances of localization errors.. Many recent SOTA method in local feature extraction \cite{superpoint,disk,CAPS,aslfeat} and matching \cite{superglue,DRCNet,SparseNCNet} can be readily integrated into HLoc framework as alternates to originally used HFNet \cite{hloc2} and NN matching.
If there were enough correctly matched local features between \textit{query} image and visual landmark map, HLoc can achieve an excellent localization performance. 
From Tab. \ref{tab:aachen}, it can be seen that extracting more local features contribute to localization, but MS detection is not constantly positive for localization \cite{d2net}.
The same local features with better feature matching methods will generate more accurate matches so they can achieve better performance \cite{superglue}.
From Tab. \ref{tab:aachen} and \ref{tab:inloc}, we can also see that accurate localization in indoor scene \cite{InLoc} is much more difficult than localization in urban scene \cite{benchmark6dof} due to repeated textures or textureless regions. In current framework of VL-MRL methods that local features are widely used as scene representation, textures of scene still be necessary.
}

\begin{table}[!t]
    \centering
    \renewcommand{\arraystretch}{1.3}
    \caption{Localization success ratio of some VL-MRL methods on InLoc Benchmark ~\cite{InLoc}, cited from \cite{patch2pix}.}
    \resizebox{0.5\textwidth}{!} {
        \begin{tabular}{l c c}
            \hline
            \hline
            \multirow{2}{*}{Method}                        & \multicolumn{2}{c}{Localized Queries (\%, 0.25$m$/0.5$m$/1.0$m$, 10$^\circ$)}                               \\
                                                           & DUC1                                                                          & DUC2                        \\
            \hline
            SuperPoint~\cite{superpoint}              & 40.4 / 58.1 / 69.7                                                            & 42.0 / 58.8 / 69.5          \\
            D2Net~\cite{d2net}                        & 38.4 / 56.1 / 71.2                                                            & 37.4 / 55.0 / 64.9          \\
            R2D2~\cite{r2d2}                          & 36.4 / 57.6 / 74.2                                                            & 45.0 / 60.3 / 67.9          \\
            \makecell[l]{SuperPoint\\~~+ SuperGlue~\cite{superglue}}        & 49.0 / {68.7} / 80.8                                                   & 53.4 / 77.1 / {82.4} \\
            SparseNCNet~\cite{SparseNCNet}                 & 41.9 / 62.1 / 72.7                                                            & 35.1 / 48.1 / 55.0          \\
            Patch2Pix \cite{patch2pix}                     & 44.4 / 66.7 / 78.3                                                            & 49.6 / 64.9 / 72.5          \\
            \makecell[l]{Patch2Pix\\~~(w.SuperGlue) \cite{patch2pix}}       & {50.0} / 68.2 / {81.8}                                          & {57.3 / 77.9 }/ 80.2 \\  
            \hline
            \hline
        \end{tabular}
    }
    \label{tab:inloc}
\end{table}

\begin{table}[!h]
    \centering
    \renewcommand{\arraystretch}{1.3}
    \caption{6 DoF localization errors of some I2P-MRL methods on KITTI Odometry Dataset \cite{KITTI}}
    \resizebox{0.5\textwidth}{!}
    {
    \begin{threeparttable}
        \begin{tabular}{lcccc}
            \hline
            \hline
            \multirow{2}{*}{Method}                     & \multicolumn{2}{c}{Mean}     & \multicolumn{2}{c}{Median}                                                           \\
                                                        & Rot.~($^\circ$)~$\downarrow$ & Transl.~(m)~$\downarrow$   & Rot.~($^\circ$)~$\downarrow$ & Transl.~(m)~$\downarrow$ \\
            \hline
            \multicolumn{5}{l}{\textbf{Geometry-based Cross-modal Localization}}                                                                                              \\
            \hline
            Caselitz \textit{et al.}~\cite{mono_i2p_vo} & 1.65                         & 0.30                       & n.a.                         & n.a.                     \\
            \hline
            \multicolumn{5}{l}{\textbf{Cross-modal Relative Pose Regression}}                                                                                                 \\
            \hline
            CMRNet~\cite{cmrnet}                        & 1.09                         & 0.26                       & 1.39                         & 0.51                     \\
            HyperMap~\cite{hypermap}                    & n.a.                         & n.a.                       & 1.42                         & 0.48                     \\
            POET~\cite{miao2023poses}                   & 0.91                         & 0.25                       & 0.79                         & 0.20                     \\
            I2PNet~\cite{I2PNet}                        & 0.74                         & 0.08                       & 0.67                         & 0.07                     \\
            \hline
            \multicolumn{5}{l}{\textbf{Cross-modal Matching-based Localization}}                                                                                              \\
            \hline
            CMRNet++~\cite{cmrnet++}                    & n.a.                         & n.a.                       & 1.46                         & 0.55                     \\
            I2D-Loc~\cite{I2DLoc}                       & n.a.                         & n.a.                       & 0.70                         & 0.18                     \\
            \hline
            \hline
        \end{tabular}
        \begin{tablenotes}
            \item[1] The measured 6 DoF localization errors include 3 DoF rotation errors (Rot.) and 3 DoF translation errors (Transl.).
        \end{tablenotes}
    \end{threeparttable}
    }
    \label{tab:i2p}
\end{table}

{
\vspace{4pt}
\setlength{\parindent}{0cm}
\textbf{I2P-MRL: }
Tab. \ref{tab:i2p} shows some cross-modal MRL methods on KITTI 00 sequence \cite{KITTI}. The learning-based I2P-MRL methods \cite{cmrnet,cmrnet++,hypermap,miao2023poses,I2PNet,I2DLoc} seem to outperform geometry-based one \cite{mono_i2p_vo} with a lot of margins.
The SOTA I2P-MRL methods can achieve an impressive localization performance with 0.08 m in translation error and 0.74\degree in rotation error \cite{I2PNet}.
However, these I2P-MRL methods are trained and tested on human-defined data that the displacement between \textit{query} image and \textit{reference} point cloud is randomly determined within a defined threshold and it is often limited. So, their performance in practical applications is still waiting for evaluation.
}

\begin{table}[htbp]
    \centering
    \renewcommand{\arraystretch}{1.3}
    \caption{3 DoF Localization errors of some HD-MRL methods.}
    \resizebox{0.5\textwidth}{!}
    {
    \begin{threeparttable}
        \begin{tabular}{lcccc}
            \hline
            \hline
            Method & Dataset & Lon. (m)~$\downarrow$ & Lat. (m)~$\downarrow$ & Heading (\degree)~$\downarrow$ \\
            \hline
            \multicolumn{5}{l}{\textbf{Traditional methods}}   \\
            \hline
            \multirow{2}{*}{Choi \textit{et al.} \cite{Choi2019Access}} & 16km highways & 0.12 & 0.18 & n.a. \\
             & 21km highways & 0.10 & 0.25 & n.a. \\
             Xiao \textit{et al.} \cite{Xiao2018TISC} & KITTI 04 \cite{KITTI} & \multicolumn{3}{c}{0.345} \\
             \multirow{2}{*}{Liao \textit{et al.} \cite{Liao2019CoarseToFineVL}} & KAIST 26 \cite{KAIST} & \multicolumn{2}{c}{0.64} & 0.90 \\
              & Virtual KITTI \cite{gaidon2016virtual} & \multicolumn{2}{c}{0.29} & 0.32 \\
             \multirow{2}{*}{MLVHM \cite{MLVHM}} & Campus & \multicolumn{3}{c}{0.21} \\
             & Urban & \multicolumn{3}{c}{0.29} \\
             \multirow{2}{*}{Wen \textit{et al.} \cite{Wen2020IV}} & Campus & \multicolumn{3}{c}{0.14} \\
              & Urban & \multicolumn{3}{c}{0.25} \\
             ROADMap \cite{RoadMap} & 22km Urban & 0.04 & 0.04 & 0.12 \\
            \multirow{2}{*}{Wang \textit{et al.} \cite{Wang2021ICRA}} & Urban canyons & 0.46 & 0.13 & 0.11 \\
             & Viaducts & 0.45 & 0.13 & 0.11 \\
            \multirow{6}{*}{TM$^3$Loc \cite{TM3Loc}} & KAIST 34 \cite{KAIST} & 0.15 & 0.05 & 0.46 \\
             & KAIST 23 \cite{KAIST} & 0.32 & 0.05 & 0.61 \\
             & KAIST 26 \cite{KAIST} & 0.16 & 0.05 & 0.48 \\
             & 3.2km Shougang Park & 0.16 & 0.07 & 2.87 \\
             & 2.0km Shougang Park & 0.23 & 0.12 & 2.76 \\
             & 1.3km Shougang Park & 0.18 & 0.06 & 2.05 \\
            \hline
            \multicolumn{5}{l}{\textbf{Learning-based methods}}                 \\
            \hline
            Zhang \textit{et al.} \cite{Zhang2023TIV} & Highway & 0.17 & 0.04 & n.a. \\
            \multirow{2}{*}{BEV-Locator \cite{bevlocator}} & nuScenes \cite{nuscenes2019} & 0.18 & 0.08 & 0.51 \\
             & Qcraft & 0.14 & 0.05 & 0.25 \\
             \multirow{2}{*}{EgoVM \cite{he2023egovm}} & nuScenes \cite{nuscenes2019} & 0.11 & 0.03 & 0.09 \\
              & self-collected dataset & 0.04 & 0.03 & 0.08 \\
            \hline
            \hline
        \end{tabular}
        \begin{tablenotes}
            \item[1] The measured 3 DoF localization errors include longitudinal (Lon.), lateral (Lat.), and Heading angle errors.
        \end{tablenotes}
        
    \end{threeparttable}
    }
    \label{tab:hd}
\end{table}

{
\vspace{4pt}
\setlength{\parindent}{0cm}
\textbf{HD-MRL: } In Tab. \ref{tab:hd}, we collect the results of some HD-MRL methods from their publications. Since the public HD map data is very few in existing dataset, researchers often collect and build their own dataset to evaluate the proposed methods. For fair and full comparison, we also list the evaluated scene/dataset for each method. It can be seen that the tightly coupling of VO and HD-MRL \cite{TM3Loc} improves the accuracy in long-term localization. The learning-based methods effectively solve the data association problem between 2D images and HD map elements, thus outperforming traditional methods, and they can localize without the requirements of online VO threads. BEV-Locator \cite{he2023egovm} estimates pose by regression whereas EgoVM \cite{he2023egovm} relies on exhaustively searching pose candidates, the latter solution achieves better performance. It should be noticed that HD map is a very compact format of map representation. In MLVHM \cite{MLVHM}, the HD maps stored in ASCII format only requires about 50 KB storage space per kilometer (km), compared to about 600 MB/km of the raw point cloud map. It is not the only case that the map size of EgoVM is 0.35 MB/km compared to 5.92 MB/km of DA4AD \cite{DA4AD}. Therefore, the compactness and accuracy of HD map makes it a popular choice for localization in AD. 
}

\begin{table}[!t]
    \caption{Localization results on Cambridge Landmarks dataset \cite{posenet}. }
    \label{tab:cambridge}
    \centering
    \renewcommand{\arraystretch}{1.3}
    \resizebox{0.5\textwidth}{!}{
        \begin{tabular}{lcccc}
            \hline
            \hline
            Method  & College  & Hospital  & Shops  & St. Mary's       \\
            \hline
            \multicolumn{5}{l}{\textbf{Localization with Geo-tagged Frame Map}}  \\
            \hline
            \multicolumn{5}{l}{\textit{Visual Place Recognition}}  \\
            \hline
            VLAD~\cite{tokyo247}                         &
            2.80,5.7\degree                              & 4.01,7.1\degree           & 1.11,7.6\degree    & 2.31,8.0\degree                     \\
            VLAD+Inter~\cite{Sattler2019UnderstandingTL} & 1.48,4.5\degree           & 2.68,4.6\degree                                      & 0.90,4.3\degree           & 1.62,6.1\degree                                          \\
            \hline
            \multicolumn{5}{l}{\textit{Relative Pose Regression}} \\
            \hline
            EssNet~\cite{EssNet}                         &
            0.76,1.9\degree                              & 1.39,2.8\degree           & 0.84,4.3\degree    & 1.32,4.7\degree                     \\
            NC-EssNet~\cite{EssNet}                      & 0.61,1.6\degree           & 0.95,2.7\degree    & 0.70,3.4\degree                                      & 1.12,3.6\degree                                                                      \\
            RelocGNN\cite{DSAC2}                         & 0.48,1.0\degree           & 1.14,2.5\degree    & 0.48,2.5\degree                                      & 1.52,3.2\degree                                                                      \\
            \hline
            \multicolumn{5}{l}{\textbf{Localization with Visual Landmark Map}}                                                                  \\
            \hline
            ActiveSearch \cite{VPS}                      & 0.42,0.6\degree           & 0.44,1.0\degree                                      & 0.12,0.40\degree          & 0.19,0.5\degree                                          \\
            InLoc\cite{InLoc}                            & 0.18,0.6\degree           & 1.2,0.6\degree     & 0.48,1.0\degree  & 0.46,0.8\degree  \\
            \hline
            \multicolumn{5}{l}{\textbf{Localization with Point Cloud Map}}   \\
            \hline
            PixLoc \cite{pixloc} & 0.14/0.2\degree & 0.16/0.3\degree & 0.05/0.2\degree & 0.10/0.3\degree \\
            \hline
            \multicolumn{5}{l}{\textbf{Localization with Learnt Implicit Map}}   \\
            \hline
            \multicolumn{5}{l}{\textit{Single Scene Absolute Pose Regression}} \\
            \hline
            PoseNet \cite{posenet}                       & 1.92,5.40\degree          & 2.31,5.38\degree   & 1.46,8.08\degree
                                                         & 2.65,8.48\degree                                                                     \\
            BayesianPN \cite{BayesianPN}                 & 1.74,4.06\degree          & 2.57,5.14                                                %
            \degree                                      & 1.25,7.54\degree          & 2.11,8.38\degree                                         \\
            LSTM-PN \cite{LSTMPN}                        & 0.99,3.65\degree          & 1.51,4.29\degree   &
            1.18,7.44\degree                             & 1.52,6.68\degree                                                                     \\
            MapNet \cite{mapnet}                         & 1.07,1.89\degree          & 1.94,3.91\degree
                                                         & 1.49,4.22\degree          & 2.00,4.53\degree                                         \\
            \hline
            \multicolumn{5}{l}{\textit{Multiple Scene Absolute Pose Regression}}                                                                \\
            \hline
            MSPN \cite{MSPN}                             & 1.73,3.65\degree          & 2.55,4.05\degree   & 2.92,7.49\degree &
            2.67/6.18\degree                                                                                                                    \\
            MS-Trans\cite{MS-Trans1}                     & 0.83,1.47\degree          & 1.81,2.39\degree   & 0.86,3.07\degree & 1.62,3.99\degree \\
            c2f-MS-Trans \cite{MS-Transformer2}          & 0.70,2.69\degree          & 1.48,2.94                                                %
            \degree                                      & 0.59,2.88\degree          & {1.14},3.88\degree                                       \\
            \hline
            \multicolumn{5}{l}{\textit{Scene Coordinate Regression}}   \\
            \hline
            {DSAC \cite{DSAC}}                           & 0.30,0.5\degree           & 0.33,0.6\degree    & 0.09,0.4\degree  &
            0.55,1.6\degree                                                                                                                     \\
            {DSAC++ \cite{Less_is_More}}                 & 0.18,0.3\degree           & 0.20,0.3\degree    & 0.06,0.3\degree  & 0.13,0.4\degree  \\
            DSAC${}^{*}$ \cite{DSAC2}                    & 0.18,0.3\degree           & 0.21,0.4\degree    & 0.05,0.3\degree  & 0.15,0.5\degree  \\
            SANet \cite{SANet} & 0.32/0.5\degree  & 0.32/0.5\degree & 0.10/0.5\degree & 0.16/0.6\degree \\
            \hline
            \multicolumn{5}{l}{\textit{Absolute Pose Regression with NeRF Enhancement}}                                                         \\
            \hline
            DFNet \cite{DFNet}                           & n.a.                      & 0.46,0.87\degree   & 0.16,0.59\degree & 0.50,1.49\degree \\
            LENS \cite{Moreau2021LENSLE}                 & 0.33,0.5\degree           & 0.44,0.9\degree    & 0.27,1.6\degree  & 0.53,1.6\degree  \\
            \hline
            \multicolumn{5}{l}{\textit{NeRF as Pose Estimator}}                                                         \\
            \hline
            NeRF-Loc \cite{liu2023nerfloc} & 0.11,0.2\degree & 0.18,0.4\degree & 0.04,0.2\degree & 0.07,0.2\degree \\
            \hline
            \hline
        \end{tabular}
    }
\end{table}

{
\vspace{4pt}
\setlength{\parindent}{0cm}
\textbf{NN-MRL: }
Then, we discuss about MRL with learnt implicit map. According to the localization results on medium-scale Cambridge Landmarks dataset \cite{posenet} in Tab. \ref{tab:cambridge} and small-scale 7Scenes dataset \cite{posenet} in Tab. \ref{tab:7scenes}, we find that the localization accuracy of APR methods is not pleased, and they performs much worse than VL-MRL methods. When enhanced by NeRF, the performances of APR methods can be significantly improved, and achieve comparable performances with VL-MRL methods.
As a surprise, we find that SCR methods work well and achieve impressive localization accuracy in both medium-scale outdoor and small-scale indoor scenes. DSAC \cite{DSAC} and its extended works \cite{Less_is_More,DSAC2} even work better than VL-MRL methods in some cases. SCR may be a promising trend in MRL research in the future. But its heavy cost of training should be considered. 
As for NeRF, it can significantly boost APR methods \cite{DFNet,DPN,Moreau2021LENSLE} and it also can work well as a pose estimator \cite{liu2023nerfloc}. We are still waiting for the usage of NeRF in large-scale and long-term localization tasks since building NeRF for unbounded scene is a much more challenging problem compared to indoor scene. 
}

{
\vspace{4pt}
\setlength{\parindent}{0cm}
\textbf{Summary:}
Finally, we compare many MRL methods with various kind of scene map on outdoor Cambridge Landmarks \cite{posenet} and indoor 7Scenes \cite{posenet}, as listed in Tab. \ref{tab:cambridge} and Tab. \ref{tab:7scenes}. In outdoor scene, the rankings of MRL methods with regard to localization accuracy should be: NeRF pose estimator \cite{liu2023nerfloc} $>$ I2P-MRL \cite{pixloc} $>$ SCR $>$ APR+NeRF $\approx$ VL-MRL $>$ RPR $>$ APR $>$ VPR. 
And in indoor scene, the rankings of MRL methods are slightly changed as: NeRF pose estimator \cite{liu2023nerfloc} $\approx$ I2P-MRL \cite{pixloc} $\approx$ SCR $\approx$ APR+NeRF $\approx$ VL-MRL $>$ RPR $>$ APR $>$ VPR. Please note that this general conclusion is drawn from the limited experimental results in this survey, there may be some isolated examples in MRL research.
}

It can be seen that in small-scale indoor scene, although APR and RPR work worse than other MRL methods, they can also achieve pleased localization performance. Given the low requirement of built map in APR and RPR, they can be used in some applications that do not need high-precision localization, such as VR. 
VPR only retrieve visually similar \textit{reference} image and regard the \textit{reference} pose as an approximated \textit{query} pose, so its localization accuracy is very limited when \textit{reference} images are few in outdoor scene. 
In indoor scene, most MRL methods with various kinds of scene map achieve comparable localization performance. But in outdoor scene, the gaps among different methods are more significant. 
Using NeRF as pose estimator like NeRF-Loc \cite{liu2023nerfloc}, I2P-MRL \cite{pixloc}, SCR, and APR+NeRF seem to be promising ways in MRL researches. Adapting NeRF-based localization to large-scale unbounded scenes, reducing the requirement of high-precision point cloud map for I2P-MRL, reducing the training cost of SCR should be explored to make the MRL methods usable.

\section{Opening Discussion}
\label{sec:discussion}

In this section, we discuss about some opening questions in MRL field, and provide our opinion. The answers of following questions are still unavailable, we only hope our personal opinion can motivate further contributions in MRL research.

\subsection{Are Explicit 3D Models Necessary in Visual Localization?}

Traditional MRL methods require an explicit 3D scene map, \textit{e.g.}, visual landmarks map in VL-MRL, point cloud map in I2P-MRL, and vectorized map in HD-MRL. An explicit 3D model of the scene appears to be necessary in the early stage of MRL research.
However, with the recent advancements in MRL, some techniques seem no longer require an explicit 3D model, such as VPR, RPR, and NN-MRL. Therefore, a question naturally arises: \textit{are explicit 3D models necessary for accurate visual localization?}

As shown in Tab. \ref{tab:cambridge} and \ref{tab:7scenes}, it can be seen that some 2D image-based methods without explicit map, like VPR and RPR, perform worse than 3D model-based method with explicit map, such as VL-MRL and I2P-MRL. However, 2D image-based methods with a learnt implicit map, \textit{i.e.}, NN-MRL, can achieve comparable or even better performances. Specially, NeRF-Loc \cite{liu2023nerfloc} has been found to localize with centimeter-level accuracy, surpassing VL-MRL and I2P-MRL by a considerable margin. Thus, it is believed that the learnable implicit map has the potential to be a superior alternative to the explicit map for accurate localisation. \textit{The explicit 3D map is not strictly necessary} in current DL age.

\begin{table*}[!h]
    \caption{Localization results on 7Scenes dataset \cite{posenet}.}
    \label{tab:7scenes}%
    \centering
    \renewcommand{\arraystretch}{1.3}
    \resizebox{1.0\textwidth}{!}{
        \begin{tabular}{lcccccccc}
            \hline
            \hline
                                                        & Chess              & Fire                & Heads              & Office             & Pumpkin             & Kitchen             & Stairs             & Average            \\
            \hline
            \multicolumn{5}{l}{\textbf{Localization with Geo-tagged Frame Map}} \\
            \hline
            \multicolumn{5}{l}{\textit{Visual Place Recognition}}                                                                                                                                                                  \\
            \hline
            VLAD~\cite{tokyo247}                        & 0.21,12.5\degree   & 0.33,13.8\degree    & 0.15,14.9\degree   & 0.28,11.2\degree   & 0.31,11.2\degree    & 0.30,11.3\degree    & 0.25,12.3\degree   & 0.26,12.46\degree  \\
            VLAD+Inter\cite{Sattler2019UnderstandingTL} & 0.18,10.0\degree   & 0.33,12.4\degree    & 0.14,14.3\degree   & 0.25,10.1\degree   & 0.26,9.4\degree     & 0.27,11.1\degree    & 0.24,14.7\degree   & 0.24,11.71\degree  \\
            \hline
            \multicolumn{5}{l}{\textit{Relative Pose Regression}}       \\
            \hline
            NN-Net~\cite{Zakaria2017ICCVW}              & 0.13,6.5\degree    & 0.26,12.7\degree    & 0.14,12.3\degree   & 0.21,7.4\degree    & 0.24,6.4\degree     & 0.24,8.0\degree     & 0.27,11.8\degree   & 0.21,9.30\degree   \\
            RelocNet\cite{RelocNet}                     & 0.12,4.1\degree    & 0.26,10.4\degree    & 0.14,10.5\degree   & 0.18,5.3\degree    & 0.26,4.2\degree     & 0.23,5.1\degree     & 0.28,7.5\degree    & 0.21,6.73\degree   \\
            EssNet~\cite{EssNet}                        & 0.13,5.1\degree    & 0.27,10.1\degree    & 0.15,9.9\degree    & 0.21,6.9\degree    & 0.22,6.1\degree     & 0.23,6.9\degree     & 0.32,11.2\degree   & 0.22,8.03\degree   \\
            NC-EssNet~\cite{EssNet}                     & 0.12,5.6\degree    & 0.26,9.6\degree     & 0.14,10.7\degree   & 0.20,6.7\degree    & 0.22,5.7\degree     & 0.22,6.3\degree     & 0.31,7.9\degree    & 0.21,7.50\degree   \\
            \hline
            
            \multicolumn{5}{l}{\textbf{Localization with Visual Landmark Map}}  \\
            \hline
            Active Search \cite{VPS}                    & 0.04,2.0\degree    & 0.03,1.5\degree     & 0.02,1.5\degree    & 0.09,3.6\degree    & 0.08,3.1\degree     & 0.07,3.4 \degree    & 0.03,2.2\degree    & 0.05,2.47\degree   \\
            InLoc\cite{InLoc}                           & 0.03,1.05\degree   & 0.03,1.07\degree    & 0.02,1.16\degree   & 0.03,1.05\degree   & 0.05,1.55\degree    & 0.04,1.31\degree    & 0.09,2.47\degree   & 0.04,1.44\degree   \\
            \hline

            \multicolumn{5}{l}{\textbf{Localization with Point Cloud Map}}  \\
            \hline
            PixLoc \cite{pixloc} & 0.02/0.8\degree & 0.02/0.7\degree & 0.01/0.8\degree & 0.03/0.8\degree & 0.04/1.2\degree & 0.03/1.2\degree & 0.05/1.3\degree & n.a. \\
            \hline

            \multicolumn{5}{l}{\textbf{Localization with Learnt Implicit Map}}  \\
            \hline
            \multicolumn{5}{l}{\textit{Single Scene Absolute Pose Regression}}   \\
            \hline
            PoseNet \cite{posenet}                      & 0.32,8.12\degree   & 0.47,14.4\degree    & 0.29,12.0\degree   & 0.48,7.68\degree   & 0.47,8.42\degree    & 0.59,8.64\degree    & 0.47,13.8\degree   & 0.44,10.44\degree  \\
            BayesianPN \cite{BayesianPN}                & 0.37,7.24\degree   & 0.43,13.7\degree    & 0.31,12.0\degree   & 0.48,8.04\degree   & 0.61,7.08\degree    & 0.58,7.54\degree    & 0.48,13.1\degree   & 0.47,9.81\degree   \\
            LSTM-PN \cite{LSTMPN}                       & 0.24,5.77\degree   & 0.34,11.9\degree    & 0.21,13.7\degree   & 0.30,8.08\degree   & 0.33,7.00\degree    & 0.37,8.83\degree    & 0.40,13.7\degree   & 0.31,9.85\degree   \\
            AtLoc\cite{Wang2019AtLocAG}                 & {0.10},4.07\degree & 0.25,11.4\degree    & 0.16,{11.8}\degree & 0.17,{5.34}\degree & 0.21,{4.37} \degree & 0.23,5.42\degree    & 0.26,10.5\degree   & 0.20,7.56\degree   \\
            \hline
            \multicolumn{5}{l}{\textit{Multiple Scene Absolute Pose Regression}}   \\
            \hline
            MSPN\cite{MSPN}                             & 0.09/4.76\degree   & 0.29/10.5\degree    & 0.16/13.1\degree   & 0.16/6.80\degree   & 0.19/5.50\degree    & {0.21}/6.61\degree  & 0.31/11.6\degree   & 0.20/7.56\degree   \\
            MS-Trans\cite{MS-Trans1}                    & 0.11,4.66\degree   & {0.24,9.60} \degree & 0.14,12.2\degree   & 0.17,5.66\degree   & 0.18,4.44\degree    & 0.17,5.94 \degree   & 0.26,8.45\degree   & 0.18,{7.28}\degree \\
            c2f-MS-Trans \cite{MS-Transformer2}         & {0.10},4.60\degree & {0.24},9.88\degree  & {0.12},12.3\degree & {0.16},5.64\degree & 0.16,4.42\degree    & {0.16},6.39\degree  & {0.25,7.76}\degree & {0.17/7.28}\degree \\
            \hline
            \multicolumn{5}{l}{\textit{Scene Coordinate Regression}}    \\
            \hline
            {DSAC \cite{DSAC}}                          & 0.02,0.7\degree    & 0.03,1.0\degree     & 0.02,1.30\degree   & 0.03,1.0\degree    & 0.05,1.30\degree    & 0.05,1.5\degree     & 1.90,49.4\degree   & 0.30,8.03\degree   \\
            {DSAC++\cite{Less_is_More}}                 & 0.02,0.5\degree    & 0.02,0.9\degree     & 0.01,0.8           & 0.03,0.7\degree    & 0.04,1.1\degree     & 0.04,1.1\degree     & 0.09,2.6\degree    & 0.04,1.10\degree   \\
            DSAC${}^*$ \cite{DSAC2}                     & 0.02,1.1\degree    & {0.02},1.0\degree   & 0.01,1.8\degree    & 0.03,1.2\degree    & 0.04,1.4\degree     & 0.03,1.7\degree     & 0.04,1.4\degree    & 0.03,1.37\degree   \\
            SANet \cite{SANet} & 0.03/0.9\degree & 0.03/1.1\degree & 0.02/1.5\degree & 0.03/1.0\degree & 0.05/1.3\degree & 0.04/1.4\degree & 0.16/4.6\degree & n.a. \\
            \hline
            \multicolumn{5}{l}{\textit{Absolute Pose Regression with NeRF Enhancement}}   \\
            \hline
            DFNet \cite{DFNet}                          & 0.04,1.48\degree   & 0.04,2.16\degree    & 0.03,1.82\degree   & 0.07,2.01\degree   & 0.09,2.26\degree    & 0.09,2.42\degree    & 0.14,3.31\degree   & 0.07,2.21\degree   \\
            Direct-PN+U \cite{DPN}             & 0.09,2.77\degree   & 0.16,4.87\degree    & 0.10,6.64\degree   & 0.17,5.04\degree   & 0.19,3.59\degree    & 0.19,4.79\degree    & 0.24,8.52\degree   & 0.16,5.17\degree   \\
            LENS \cite{Moreau2021LENSLE}                & 0.03,1.3\degree    & 0.10,3.7\degree     & 0.07,5.8\degree    & 0.07,1.9\degree    & 0.08,2.2\degree     & 0.09,2.2\degree     & 0.14,3.6\degree    & 0.08,3.0\degree    \\
            \hline
            \multicolumn{5}{l}{\textit{NeRF as Pose Estimator}}                                                         \\
            \hline
            NeRF-Loc \cite{liu2023nerfloc} & 0.02/1.1\degree & 0.02/1.1\degree & 0.01/1.9\degree & 0.02/1.1\degree & 0.03/1.3\degree & 0.03/1.5\degree & 0.03/1.3\degree & n.a. \\
            \hline
            \hline
        \end{tabular}
    }
\end{table*}

\subsection{How to Select Specific Localization Method?}

Hundreds of MRL researches are proposed each year, relying on various kinds of scene map and achieving differing levels of localization accuracy. Therefore, one may consider \textit{how to choose the appropriate MRL methods} for a specific autonomous task?

As shown in the experimental results in Sec. \ref{sec:sota}, the utilized scene map of MRL usually determines the localization accuracy that a particular kind of MRL can achieve:
1) VPR methods estimate current pose by retrieving historical posed images. Their localization accuracy is limited, but they solely require a 2D image map and can obtain \textit{reference} images with overlapped view-of-field. As a result, they are often utilized as a coarse step in hierarchical localization framework \cite{hloc1,hloc2} to acquire co-visible \textit{query-reference} image pairs. 2) With a slightly better localization accuracy, RPR and APR methods can be employed in some applications that do not necessitate high-precision poses but call for a low-cost and light-weight scene map. 3) VL-MRL and I2P-MRL methods require an offline mapping stage to obtain 3D scene map, \textit{i.e.}, visual landmarks map or point cloud map, so their cost of mapping is significantly higher than previously mentioned methods. With these precise and explicit 3D scene map, these two kind of methods achieve outstanding localization performance under most circumstances. However, when aiming for autonomy in large-scale scenarios, these two kinds of scene map require significant storage space, making them difficult to deploy on resource-limited platform, like intelligent vehicles. 4) HD map is the most light and compact form of scene map, so HD-MRL methods are widely applied in AD task. But HD Map is limited to human-made scenes with semantic map elements, and it is not suitable for natural scenes.  5) Recently developed NN-MRL methods utilize DNN model to represent scenes, which have demonstrated impressive success in small-scale indoor scene and medium-scale urban scenes. However, the efficacy of NN-MRL has yet to be fully evaluated in unbounded scenes. 

In summary, we should \textit{choose specific MRL methods based on its required localization accuracy, supported platform, and types of applied scenarios} (including textures, scene size, \textit{etc.}).

\subsection{What's Next for Visual Localization?}

Developed over several decades, MRL continues to attract growing attention from both academia and industry. We would like to discuss about an interesting and instructive problem: \textit{"What are the future trends of visual localization research?"}

{
\vspace{4pt}
\setlength{\parindent}{0cm}
\textbf{End-to-end pipeline: }
The robustness against environmental changes should be well considered by MRL methods for long-term autonomy. The current \textit{query} image often has different visual conditions with \textit{reference} data stored in the scene map. Most of existing solutions rely on image feature algorithms \cite{sift,superpoint} to attain condition-invariant image representation. 
However, these features often fail to achieve comparable performance to that on feature matching tasks. 
For instance, the matching accuracy of D2-Net \cite{d2net} is inferior to SuperPoint \cite{superpoint} on HPatches benchmark \cite{hpatches1,hpatches2}, but when it comes to localization benchmark \cite{benchmark6dof}, D2-Net \cite{d2net} yields better results than SuperPoint \cite{superpoint}. 
This discrepancy between feature matching and localization might be attributed to the limitations of human-defined supervision on features. 
Some recent approaches suggest building the whole localization pipeline into an E2E manner, incorporating feature extraction, matching, and pose estimation, and training such localization models by supervising them on estimated poses \cite{I2PNet,I2DLoc,Stumberg2019GNNetTG,pixloc}.
Such pose supervision can facilitate the learning of involved geometry priors in MRL and enhance the global consistency of feature matching. 
This kind of image feature are specifically learned for MRL, enabling the closure of the gap between feature matching and localization, thus becoming a more reasonable solution for representation in MRL research. 
It is pertinent to note that HD-MRL is encountered with similar challenges alongside matching-based localization. 
Since semantic instance-level matching is sparse and difficult, some E2E solutions prefer to make models extract and match high-dimensional features for both images and vectorized map elements, and then estimate poses, leading to improved localization performances \cite{bevlocator,he2023egovm}. So, we believe E2E localization system will be a promising trend.
}

{
\vspace{4pt}
\setlength{\parindent}{0cm}
\textbf{Resource-friendly MRL: }
To achieve real-time ego-pose estimation in large-scale scenarios during long-term exploration, we must take into account the storage demands of scene map and the running efficiency of localization algorithms. 
For example, high-level autonomous vehicles in unbounded driving areas generally depend on vectorized HD map, rather than visual landmarks map and point cloud map that are commonly used in robotic applications within confined scenes \cite{TM3Loc,MLVHM}. 
In the field of robotic localization, several recommendations are available to lighten scene map, \textit{e.g.}, SceneSqueezer \cite{SceneSqueezer}, NeuMap \cite{neumap}, and HyperMap \cite{hypermap}, which significantly decrease the map size while ensuring localization accuracy.
There is no doubt that MRL will prefer a lightweight and compact scene map in the future.
Aiming at high efficiency, MRL methods are commonly built in coarse-to-fine framework where VPR is applied to coarsely estimate poses \cite{hloc1,hloc2}. Recent advances enable us to employ NN-MRL solutions as a coarse step, such as APR in LATITUDE \cite{LATITUDE}. Additionally, improving the efficiency of key components in pipelines is also essential. ALIKE \cite{alike} efficiently extracts local features while LightGlue \cite{lindenberger2023lightglue} achieves high-speed matching. 
Efforts to reduce computational cost of MRL methods should be encouraged, as this will stimulate the proposal of more practical MRL solutions for resource-limited platforms.
}

{
\vspace{4pt}
\setlength{\parindent}{0cm}
\textbf{MRL with new kind of map: }
In recent years, numerous MRL methods that use new types of scene map have been created.
For example, MeshLoc \cite{meshloc} employs a dense mesh-based map rather than visual landmarks map and achieves superior performance. Similarly, NN-MRL methods utilize implicit neural network-based scene map as a substitute for traditional explicit 3D structure-based scene map. Proposing new representation formats of scene map is still an area of research that requires further exploration in MRL. 
The ideal scene map should be light-weight, compact, and easy to build and deploy. 
Moreover, it ought to provide comprehensive information for autonomy.
The reference data for localization in the scene map should be sufficiently robust against potential changes in real-world scene. 
Current scene map is challenging to accomplish all of these criteria, so we believe it is necessary to develop new kind of scene map for MRL, which will bring about impressive improvement to MRL researches. 
}

{
\vspace{4pt}
\setlength{\parindent}{0cm}
\textbf{Multiple sensor fusion for localization: }
We have to acknowledge a fact that visual information is sensitive to environmental interference, so visual-only localization is hard to guarantee a stable performance all the time. As a solution, in practical applications, researches usually combine MRL with other sensor-based localization systems, such as IMU, GNSS, and wheel encoder, to couple multi-sensor-based localization results in loose or tight manner. 
For example, in TM${}^{3}$Loc \cite{TM3Loc}, the authors tightly coupled HD-MRL with visual-inertial odometry to achieve accurate and consistent ego-pose estimation even when the map elements are insufficient to support localization. 
Although the multi-sensor fusion strategy has been well studied in computer vision tasks \cite{SGFNet} and SLAM researches \cite{orbslam2,orbslam3} for many decades, we believe that the strategy is still a promising area for exploration due to its value in practical applications. 
}

\section{Conclusions}
\label{sec:conclusion}

In this survey, we formulate MRL method as a interaction procedure between \textit{query} image and scene map where poses are estimated, and then we systemically review MRL methods based on the representation format of utilized scene map, that is, MRL using geo-tagged frame map, visual landmark map, point cloud map, vectorized HD map, and learnt implicit map. Each kind of MRL methods and the related component of them are fully reviewed. Also, we provide a review about evaluation of MRL methods, and draw some conclusions based on the evaluation results of typical algorithms. We provide some opening problem in this field and give our personal opinions. Finally, as a continuous contribution to the community, we list the reviewed papers on the website so that researches can readily find the best-matched MRL method based on their interests.






\vfill

\end{document}